\documentclass[10pt,twocolumn,letterpaper]{article}

\usepackage[pagenumbers]{cvpr}
\usepackage{booktabs}
\usepackage{xcolor}
\usepackage{multirow}
\usepackage{makecell}
\usepackage{colortbl}
\usepackage{adjustbox}
\usepackage{xcolor}
\usepackage[normalem]{ulem}

\usepackage{algorithm}
\usepackage{algpseudocode}
\usepackage{graphicx}
\usepackage{capt-of}
\usepackage{xcolor}
\usepackage{amsmath}
\usepackage{amssymb}

\usepackage{tabularx}
\usepackage{dblfloatfix}
\renewcommand{\topfraction}{0.85}
\renewcommand{\bottomfraction}{0.5}
\renewcommand{\textfraction}{0.1}
\renewcommand{\floatpagefraction}{0.8}
\renewcommand{\dbltopfraction}{0.85}
\renewcommand{\dblfloatpagefraction}{0.8}
\definecolor{oursbg}{gray}{0.92}
\definecolor{gaingreen}{RGB}{0,128,64}
\definecolor{oraclegray}{gray}{0.5}
\newcommand{\gain}[1]{\textsubscript{\,\textcolor{gaingreen}{#1}}}
\providecommand{\loss}[1]{\textsubscript{\,\textcolor{oraclegray}{#1}}}

\definecolor{cvprblue}{rgb}{0.21,0.49,0.74}
\usepackage[pagebackref,breaklinks,colorlinks,allcolors=cvprblue]{hyperref}
\def\confName{CVPR}
\def\confYear{2027}

\title{Describe, Don't Decide: Correcting Answer Bias in Multimodal LLMs}

\author{Shuwei Wu\qquad Longin Jan Latecki\thanks{Corresponding author.}\\
Temple University\qquad{\tt\small \{shuwei.wu, latecki\}@temple.edu}
}

\begin{document}

\twocolumn[{%
\renewcommand\twocolumn[1][]{#1}%
\maketitle
\vspace{-2.5em}%
\begin{center}
  \includegraphics[width=0.94\textwidth]{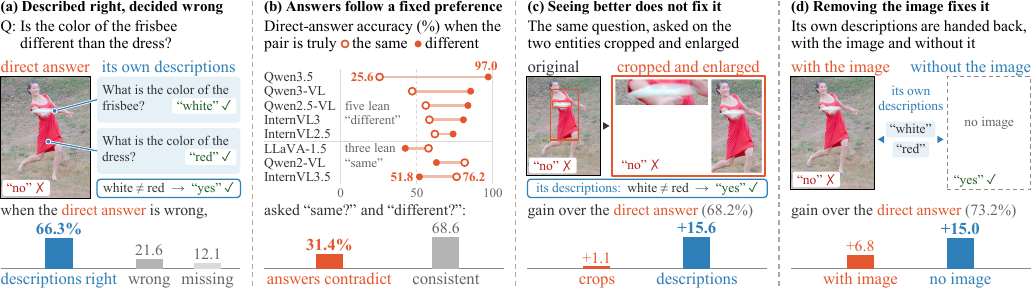}
\captionof{figure}{\textbf{The answer bias enters when the model decides with the image in view.}
  Eight MLLMs on 670 held-out GQA questions; bars are averages over the models, and the examples are from InternVL3.
  Orange: decided with the image in view; blue: from the model's own descriptions.
  (a) Asked directly, the model answers ``no'' (wrong); asked about one entity at a time, it describes both correctly, and comparing the two gives ``yes''.
  Bars: the descriptions give the right answer for 66.3\% of the wrong direct answers; missing: an entity gets no description.
  (b) Each row is a model; the hollow ($\circ$) and filled ($\bullet$) dots are its accuracy on pairs that are truly the same and truly different, and with no preference the two would coincide.
  Bars: when ``same?'' is flipped to ``different?'', the answer should flip too, but on 31.4\% of the questions it stays the same.
  (c) Given the two entities cropped by their annotated boxes and enlarged instead of the image, the model still answers ``no''; accuracy rises by only 1.1\%, against 15.6\% from comparing the descriptions.
  (d) Handing the descriptions back raises accuracy by 6.8\% with the image and by 15.0\% without it.}
  \label{fig:findings}
\end{center}%
\vspace{-0.5em}%
}]
\begin{abstract}
Comparing two entities in an image, multimodal LLMs (MLLMs) often answer wrongly.
We find that the answer bias enters not when the model sees, but when it decides with the image in view: the model's own descriptions of the two entities often already hold the right answer.
We therefore propose D\textsuperscript{3} (\textbf{D}escribe, \textbf{D}on't \textbf{D}ecide): the model describes each entity alone, decides from these descriptions without the image, and merges this decision with its direct answer.
Without any training, D\textsuperscript{3} improves all 79 model--benchmark pairs across ten benchmarks and eight 7--9B MLLMs.
On held-out GQA comparison questions, the mean accuracy of these models rises from 72.4\% to 83.2\% (up to $+$21.5\% per model), higher than that of Qwen2.5-VL-32B answering directly (82.8\%).
Code will be available at \url{https://github.com/paul-swu/describe-dont-decide}.
\par\vspace{-1em}
\end{abstract}

\section{Introduction}
\label{sec:intro}
Multimodal large language models (MLLMs) often answer wrongly about what they have already seen.
A comparison question asks how two entities in an image relate in one aspect: for example, whether they have the same color or material, which one is further left, or which group is larger.
The answer depends only on what each of the two entities is like, but models often get it wrong~\cite{tong2024mmvp,kamath2023whatsup}.
Their answers are also biased: they shift with language priors, the order of the options, and the wording of the question~\cite{guan2024hallusionbench,zong2024fool,fan2026tinted}.

To fix these errors, most existing work acts on the observation side: it adds marks to the image, highlights the relevant regions, or adjusts the model's internal attention~\cite{yang2023som,yu2024api,liu2024pai,chen2025adaptvis}.
Other work repairs the decision itself, for example with contrastive decoding, output calibration, or step-by-step reasoning~\cite{leng2024vcd,zhang2025debiasing,wei2022chain}.
Either way, the decision is still made by the model with the image in view.
Whether the bias enters at this step has not been tested.

We split the model's direct answer to the original question into two steps, seeing and deciding, and test step by step where the bias enters (Sec.~\ref{sec:findings}, Fig.~\ref{fig:findings}).
We make four findings:
(1) the model's own descriptions already hold the answer;
(2) the answer follows a fixed preference;
(3) seeing the entities better does not fix the answer;
(4) handing back the descriptions works only without the image.
Taken together, the bias does not enter when the model sees, but when it decides with the image in view.
The model needs the image to describe the two entities, but it should not look at the image when it decides.

We therefore propose D\textsuperscript{3} (\textbf{D}escribe, \textbf{D}on't \textbf{D}ecide), which describes with the image in view and decides without it.
The model first describes each entity on its own, with the image but without the original question.
It then decides from these two descriptions alone, without the image.
Finally, this decision is merged with the direct answer, and questions without a description keep the direct answer.
D\textsuperscript{3} needs no training, and by default the same frozen model does every step.
A description can be an attribute value, a box, or a number, and it can also come from another MLLM or a geometric tool.

We evaluate eight MLLMs with 7--9B parameters on ten benchmarks that cover attribute, count, and position comparisons, and all 79 model--benchmark pairs gain.
On held-out GQA comparison questions, the mean accuracy of the eight models rises from 72.4\% to 83.2\% (+10.8\%).
With D\textsuperscript{3}, these 7--9B models reach on average the accuracy of Qwen2.5-VL-32B when it answers directly from the image (83.2\% vs.\ 82.8\%).

\section{Where Does the Answer Bias Enter?}
\label{sec:findings}
When a model answers a comparison question directly, it looks at the image and decides in a single step, so a wrong answer does not show where it went wrong.
But where it went wrong determines what to fix.
We therefore take this step apart in this section and check it one piece at a time.
Our experiments use 670 held-out GQA questions, one per image, each asking whether two entities have the same color or material, and eight MLLMs with 7--9B parameters (Sec.~\ref{sec:exp_setup}).
Each model answers the original question with the image in view; we call this its \textit{direct answer}.
The most direct explanation for a wrong direct answer is that the model did not see the two entities clearly.
So we first ask: does the model see the color or material of each entity?

\paragraph{Finding 1: The model's own descriptions already hold the answer.}
To find out, we ask about one entity at a time, e.g., ``What is the color of the frisbee?'', without the original question, and the model answers in one or two words.
These two answers are the model's own \textit{descriptions}.
Comparing them by rule, with equal values meaning ``same'' and unequal values ``different'', gives the \emph{description result}.
In Fig.~\ref{fig:findings}a, InternVL3's direct answer is ``no'', but it describes the frisbee as white and the dress as red.
Seven of the eight models give this wrong direct answer, and all eight give these two descriptions.
This is not a rare case: for 66.3\% of the 1{,}679 wrong direct answers of the eight models, the description result is right (Fig.~\ref{fig:findings}a).
Over all 670 questions, the probability of ``yes'' computed from the two descriptions separates the questions whose answer is yes from the rest with an AUROC of 0.93; the direct answer's probability reaches 0.75 (0.5 is chance).
So most wrong direct answers are not failures to see.
What the model knows and what it answers are different things: here its answer does not follow what it sees.
We next ask what the answer follows instead.

\paragraph{Finding 2: The answer follows a fixed preference.}
If the answer followed nothing in particular, wrong direct answers would be occasional mistakes, about equally frequent on questions whose two entities are in fact the same (\emph{same pairs}) and on those whose entities differ (\emph{different pairs}).
They are not (Fig.~\ref{fig:findings}b).
Qwen3.5 is right on 97.0\% of different pairs but on only 25.6\% of same pairs: it mostly answers ``different''.
InternVL3.5 does the opposite, with 76.2\% on same pairs and only 51.8\% on different pairs.
Each model has its own preference: five of the eight lean toward ``different'' and three toward ``same'', so no single shift of the answers toward one side can correct all of them.
The models also disagree with each other: two models give the same direct answer on only 66.6\% of the questions on average, while their description results agree on 90.8\%, and this holds for all 28 pairs of models (Supp.).
Overall accuracy hides this preference and can even reward it: three quarters of the questions are different pairs, and the most accurate of the eight models, Qwen3.5 (79.1\%), has the strongest preference.

Moreover, the answer changes when only the wording changes.
For the same pair, we ask once whether the two entities have the same color and once whether they have different colors; the two answers should be opposites.
On 31.4\% of the questions, averaged over the models, they contradict each other (Fig.~\ref{fig:findings}b).
The image is the same but the answer changes, so the answer follows a preference of the model, not what the image shows; this is the answer bias. It also explains Finding~1: when the model describes one entity at a time, there is no ``same'' or ``different'' to choose, so the preference has nothing to act on.

However, most existing remedies act on the observation side: some add marks to the image~\cite{yang2023som,cai2024vipllava}, some highlight the relevant regions~\cite{yu2024api,an2025agla}, and some adjust the model's internal attention~\cite{liu2024pai,kang2025var,yin2025clearsight,chen2025adaptvis}.
The methods differ, but they rest on the same assumption: the model answers wrongly because it does not see the entities well enough. The first two findings suggest otherwise: in most wrong direct answers, the model's own descriptions already give the right answer (Finding 1), and the direct answer follows a fixed preference (Finding 2).
But this evidence is still indirect; we therefore intervene on the observation side directly and measure how much seeing the entities better helps.

\paragraph{Finding 3: Seeing the entities better does not fix the answer.}
If this assumption held, showing the entities clearly should fix the answer.
We give the model the best view available: for 600 of the 670 questions, both entities have annotated boxes in the GQA scene graphs; we crop the entities by these boxes, enlarge them, show them side by side, and ask the original question.
These observation-side remedies cannot locate the entities better than this.
The model sees only the relevant regions, and these regions come from oracle annotations.
Surprisingly, in Fig.~\ref{fig:findings}c, the model still answers ``no''.
Over the eight models, mean accuracy on these questions moves from 68.2\% to 69.3\%, a gain of only 1.1\%.
Marking the two entities with boxes on the original image instead lowers accuracy by 5.3\%.

We now keep the image unchanged and change how the decision is made: we use the description result from Finding~1.
Accuracy rises to 83.8\%, and all eight models gain; accuracy on same pairs also rises, from 56.6\% to 69.9\% on average.
We put the two results side by side (Fig.~\ref{fig:findings}c): with the same way of deciding, a clearer image changes accuracy by 1.1\%; with the same image, a different way of deciding changes it by 15.6\%.
A clearer image helps so little because the original image already shows the model enough: even the descriptions, when made on the crops instead, raise the description result by only another 2.3\%.
So the problem is not what the model sees, but how it decides.
If the answer is already in the model's own descriptions, why not simply hand them back when it decides?

\paragraph{Finding 4: Handing back the descriptions works only without the image.}
If the model only lacked this information when it decided, it should follow its descriptions once they are handed back.
The most natural way is to write its two descriptions into the prompt and ask the original question again with the original image; the prompt gains one sentence: ``Earlier, you described the image as follows: the color of the frisbee is white; the color of the dress is red.'' Yet in Fig.~\ref{fig:findings}d, the model reads this sentence, looks at the image, and still answers ``no''.
To see how general this is, we count the questions for which the model gives both descriptions, over seven models (LLaVA-1.5 excluded; Supp.).
On these questions, the direct answer (73.2\%) and the description result (88.1\%) are 15\% apart, and handing the descriptions back with the image recovers less than half of this gap: the model gains 6.8\% (Fig.~\ref{fig:findings}d) and stays 8.2\% below the description result.

The gap remains because the model does not always follow its descriptions, and it follows them selectively: 92.2\% of the time when they agree with its direct answer, but only 78.7\% when they disagree (means over the seven models).
In other words, with the image in view, the model simply answers the original question again and gives less weight to the descriptions that do not match its answer.
Yet the descriptions can correct the answer only where they disagree with it.
Like seeing the entities better (Finding 3), handing the descriptions back is not enough. In both cases, the model still decides with the image in view.

We therefore take the image away, and the prompt contains only the two descriptions and the original question (prompts and controls in the Supp.).
Now the model follows its own descriptions 99.7\% of the time.
The description result disagrees with the direct answer in 1{,}054 cases (summed over the seven models).
With the image in view, the model fails to follow the descriptions in 206 of them; without it, in only 2.
Accuracy reaches 88.1\%, the same as the description result, and the 15\% gap is recovered (Fig.~\ref{fig:findings}d).
The model therefore lacks neither information nor the ability to compare.
What fixes the answer is not giving the model more, but giving it one thing less when it decides: the image.

The four findings point to the same place: the answer bias enters when the model decides with the image in view.
The image plays two roles here.
Describing the entities needs it, and the original image is enough (Findings~1 and~3); but whenever it is present when the model decides, the preference comes back with it (Findings~2 and~4).
D\textsuperscript{3} keeps only the first role: it looks at the image when describing and not when deciding.
What remains is how to use the direct answer.
Summed over the eight models, the direct answer and the description result differ in 1{,}396 cases, and the direct answer is right in only 20.3\% of them, so it should not get more weight than the description result.
But for questions without a description, only the direct answer is available, so it cannot be dropped either.
This gives the three steps of D\textsuperscript{3}: the model describes each entity alone; it decides only from these descriptions, without the image; and this decision is merged with the direct answer at equal weight (Sec.~\ref{sec:method}).

\section{D\textsuperscript{3}: Describe, Don't Decide}
\label{sec:method}
\begin{figure*}[t]
  \centering
  \includegraphics[width=\linewidth]{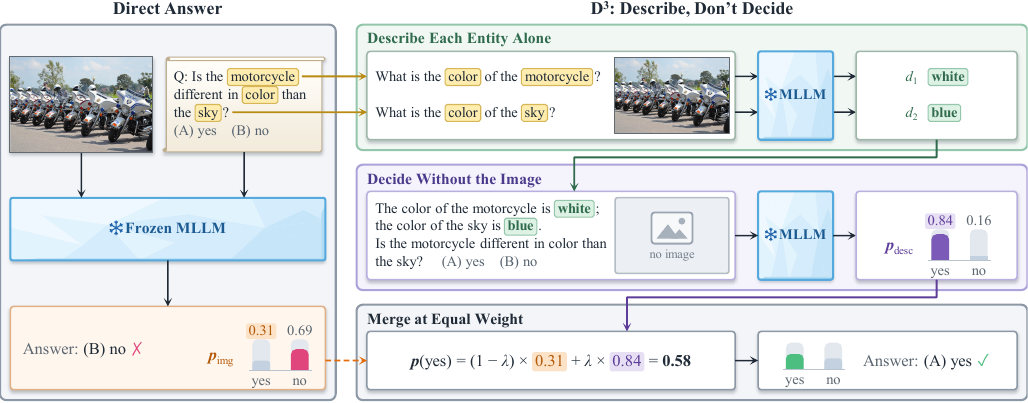}
\caption{\textbf{D\textsuperscript{3} describes with the image in view and decides without it.}
  Left: the direct answer. Given the image and the original question, the MLLM answers wrongly; $p_{\mathrm{img}}$ (orange) is its probability for each option.
  Right: the three steps of D\textsuperscript{3}; in the first two, the MLLM takes the box on its left and returns the one on its right.
  Describe Each Entity Alone: the two entities and the aspect (yellow) are read from the question, and each entity is described in its own call, with the image but not the original question, giving $d_1$ and $d_2$ (green).
  Decide Without the Image: given only $d_1$, $d_2$, and the original question (the text-only call), the MLLM gives the description-based decision $p_{\mathrm{desc}}$ (purple).
  Merge at Equal Weight: the output is the option $y$ with the larger $(1-\lambda)\,p_{\mathrm{img}}(y) + \lambda\,p_{\mathrm{desc}}(y)$, where $\lambda = \tfrac{1}{2}$ (Eq.~\ref{eq:merge}).
  The snowflake marks the frozen MLLM, the same in every call. Example: Qwen2.5-VL on one GQA question.}
  \label{fig:method}
\end{figure*}

\subsection{Overview}
\label{sec:overview}
To give a direct answer, the model looks at the image and decides in a single step, and this is where the answer bias enters (Sec.~\ref{sec:findings}).
D\textsuperscript{3} (Fig.~\ref{fig:method}) separates looking from deciding.
It describes each entity with the image in view but without the original question (Sec.~\ref{sec:describe}), decides from these descriptions without the image (Sec.~\ref{sec:decide}), and merges this decision with the direct answer at equal weight (Sec.~\ref{sec:merge}).
So no call that D\textsuperscript{3} adds is given both the image and the original question.
The direct answer is kept, and it is the output for questions where no decision can be made from the descriptions.
The model is frozen, and nothing is trained.

\subsection{Setup}
\label{sec:prelim}
Let $M$ be a frozen MLLM, $I$ an image, and $Q$ the original question.
Here $Q$ is a comparison question with answer options: it asks how two entities relate in one aspect, where an entity is an object, or a kind of object, named in $Q$. Questions in parts of What'sUp and CV-Bench name one or three entities; they are handled in the same way, with one description per entity (Supp.).
For the question in Fig.~\ref{fig:method}, ``Is the motorcycle different in color than the sky?'', the two entities are the motorcycle and the sky, and the aspect is color (they are read from $Q$ as described under Implementation).
The direct answer is the answer of $M$ given $I$ and $Q$; we write its probability for answer option $y$ as $p_{\mathrm{img}}(y) = P_M(y \mid I, Q)$.

\subsection{Describe Each Entity Alone}
\label{sec:describe}
Unlike the direct answer, the descriptions the model gives when asked about one entity at a time already hold the answer (Sec.~\ref{sec:findings}).
D\textsuperscript{3} therefore starts by describing each entity alone.
Each entity in $Q$ is described by a \emph{describer}.
Given the whole image $I$, the entity, and the aspect, the describer returns a short description $d_i$: a one- or two-word attribute value, a box, or a number.
In Fig.~\ref{fig:method}, the two descriptions are ``white'' and ``blue''.
By default, the describer is $M$ itself.
The describer is given neither $Q$ nor the other entity.

\subsection{Decide Without the Image}
\label{sec:decide}
The two descriptions are not yet an answer to $Q$; a decision still has to be made from them.
If this decision is made with the image in view, the model simply answers the original question again and follows the two descriptions only selectively (Sec.~\ref{sec:findings}).
D\textsuperscript{3} therefore makes this decision without the image: it gives $M$ only $Q$ and the two descriptions, and gets $p_{\mathrm{desc}}(y) = P_M(y \mid Q, d_1, d_2)$.
We call this the \emph{text-only call}, and $p_{\mathrm{desc}}$ the \emph{description-based decision}.
If the answer follows exactly from the values of the two descriptions, a fixed rule can replace the text-only call, and $p_{\mathrm{desc}}$ is then 0 or 1.
The description result of Sec.~\ref{sec:findings} uses such a rule: equal attribute values mean ``same'', and unequal values mean ``different''.
With either the text-only call or the rule, the image can affect the description-based decision only through the two descriptions.

\subsection{Merge at Equal Weight}
\label{sec:merge}
Every question has a direct answer $p_{\mathrm{img}}$, but not every question has a description-based decision $p_{\mathrm{desc}}$.
There is no $p_{\mathrm{desc}}$ when no decision can be made from the descriptions, which happens in three cases: (i) $Q$ is not a question as defined in Sec.~\ref{sec:prelim}, or its entities cannot be read from it; (ii) the describer gives no description for one of the entities; (iii) the rule makes no decision for the two descriptions, e.g., when the centers of two boxes are too close.
Both $p_{\mathrm{img}}$ and $p_{\mathrm{desc}}$ answer the same question: $p_{\mathrm{img}}$ is made with the image in view, and $p_{\mathrm{desc}}$ from the two descriptions alone. When there is a $p_{\mathrm{desc}}$, D\textsuperscript{3} merges the two with a weight $\lambda$ on the description-based decision,
\begin{equation}
  p(y) = (1-\lambda)\, p_{\mathrm{img}}(y) + \lambda\, p_{\mathrm{desc}}(y),
  \label{eq:merge}
\end{equation}
and when there is none, $p(y) = p_{\mathrm{img}}(y)$.
Since the direct answer should not get more weight than the description-based decision (Sec.~\ref{sec:findings}), D\textsuperscript{3} gives it the largest weight that this allows, $\lambda = \tfrac{1}{2}$, so the two are merged at equal weight.
The same $\lambda$ is used for all models and benchmarks and was fixed before the held-out evaluation (the Supp.\ describes how it was set); other values of $\lambda$ between $\tfrac{1}{2}$ and 1 change the results little on held-out GQA (Sec.~\ref{sec:exp_gain}).
D\textsuperscript{3} outputs the option with the largest $p(y)$.
So D\textsuperscript{3} can change the output only for questions that have a $p_{\mathrm{desc}}$; for all other questions, its output is the direct answer.

\begin{table}[t]
  \centering
  \caption{\textbf{Versions of D\textsuperscript{3} used in this paper.}
  Each row gives what a description is and the rule that decides from the two descriptions; on attribute values, the text-only call can be used instead of the rule.
  The merge (Eq.~\ref{eq:merge}) is the same in every row.}
  \label{tab:instances}
  \footnotesize
  \setlength{\tabcolsep}{4pt}
  \renewcommand{\arraystretch}{1.1}
  \begin{tabularx}{\linewidth}{@{}l>{\raggedright\arraybackslash}p{0.28\linewidth}>{\raggedright\arraybackslash}X@{}}
    \toprule
    Comparison & Description & Rule \\
    \midrule
    \multicolumn{3}{@{}l}{\textit{The model (default) or a second MLLM describes}} \\
    Attribute & a value, e.g., ``white'' & equal or not; on ARO, which option the two values match \\
    Position & a box & compare the two box centers \\
             & a horizontal position (0--100) & compare the two positions \\
    Count & an integer, or one box per object & compare the two counts \\
    \midrule
    \multicolumn{3}{@{}l}{\textit{A geometric tool describes}} \\
    Depth, distance & a depth value or a 3D point & compare the depths or the distances \\
    \bottomrule
  \end{tabularx}
\end{table}

\paragraph{Implementation.}
The two entities and the aspect are read from the text of $Q$ with a fixed set of sentence patterns, without the image; if they cannot be read, the output of D\textsuperscript{3} is the direct answer.
Every call is a new conversation and carries no context from any other call.
Each entity gets one describe call, and the prompt allows the describer to answer ``cannot tell''; if the describer answers so, or if the needed value cannot be taken from its answer, we count this as no description.
On ARO, where the two options give the same two attribute words to the two entities in swapped order, the describer chooses between these two words. The direct answer and the text-only call each generate a single token; the probabilities that the model gives to the answer options at this token, scaled to sum to one over the options, are $p_{\mathrm{img}}$ and $p_{\mathrm{desc}}$.
In total, D\textsuperscript{3} adds one describe call with a short output per entity, plus one text-only call if the decision is not made by the rule.
The prompts, the sentence patterns, and the rules are the same for all models and are given in full in the Supp.

\section{Experiments}
\label{sec:exp}

\begin{table*}[!t]
\centering
\setlength{\heavyrulewidth}{1.1pt}\setlength{\lightrulewidth}{0.6pt}\setlength{\cmidrulewidth}{0.4pt}
\caption{\textbf{D\textsuperscript{3} improves every model on every benchmark.}
Accuracy (\%) of the direct answer and of D\textsuperscript{3} (gray), with the gain over the direct answer as subscripts; the number under each benchmark is its number of questions.
GQA is the held-out set, GQA-testdev holds the attribute comparisons of GQA testdev, and What'sUp is the mean over its six subsets.
Each model uses its own descriptions, except: $^{\dagger}$\,boxes from Qwen2.5-VL; $^{\ddagger}$\,boxes and counts from Qwen2.5-VL, and depth values and 3D points from geometric tools.
Avg.: mean over the models; ``--'': not evaluated.}
\label{tab:main}
\renewcommand{\arraystretch}{1.1}
\setlength{\tabcolsep}{3pt}
\begin{adjustbox}{max width=\textwidth}
\begin{tabular}{@{}l cccccccccc@{}}
\toprule
Model & GQA & GQA-testdev & VAW-Nat & VAW-Pairs & ARO & CLEVR-Attr & CLEVR-Count & What'sUp$^{\dagger}$ & EmbSpatial$^{\dagger}$ & CV-Bench$^{\ddagger}$ \\
 & {\scriptsize 957} & {\scriptsize 326} & {\scriptsize 554} & {\scriptsize 1,200} & {\scriptsize 9,273} & {\scriptsize 1,086} & {\scriptsize 1,200} & {\scriptsize 4,341} & {\scriptsize 3,275} & {\scriptsize 2,374} \\
\midrule
LLaVA-1.5-7B & 55.9 & 63.2 & 50.7 & 50.6 & 68.3 & -- & 53.1 & 58.0 & 46.8 & 61.1 \\
\rowcolor{oursbg}\,+\,\textbf{D\textsuperscript{3}} & \textbf{77.4}\gain{+21.5} & \textbf{75.2}\gain{+12.0} & \textbf{72.9}\gain{+22.2} & \textbf{66.3}\gain{+15.8} & \textbf{77.1}\gain{+8.8} & -- & \textbf{59.6}\gain{+6.5} & \textbf{95.5}\gain{+37.6} & \textbf{75.0}\gain{+28.2} & \textbf{84.7}\gain{+23.6} \\
Qwen2-VL-7B & 75.8 & 72.7 & 63.5 & 54.8 & 93.4 & 98.7 & 92.1 & 93.5 & 64.6 & 75.5 \\
\rowcolor{oursbg}\,+\,\textbf{D\textsuperscript{3}} & \textbf{88.2}\gain{+12.4} & \textbf{78.5}\gain{+5.8} & \textbf{76.2}\gain{+12.6} & \textbf{72.2}\gain{+17.3} & \textbf{95.1}\gain{+1.7} & \textbf{100.0}\gain{+1.3} & \textbf{99.6}\gain{+7.5} & \textbf{98.3}\gain{+4.8} & \textbf{76.2}\gain{+11.6} & \textbf{86.4}\gain{+11.0} \\
Qwen2.5-VL-7B & 82.2 & 77.9 & 60.5 & 52.9 & 95.6 & 97.0 & 82.8 & 91.4 & 69.6 & 80.1 \\
\rowcolor{oursbg}\,+\,\textbf{D\textsuperscript{3}} & \textbf{84.5}\gain{+2.3} & \textbf{78.5}\gain{+0.6} & \textbf{65.9}\gain{+5.4} & \textbf{62.7}\gain{+9.8} & \textbf{96.2}\gain{+0.5} & \textbf{99.6}\gain{+2.7} & \textbf{99.3}\gain{+16.5} & \textbf{97.7}\gain{+6.3} & \textbf{78.1}\gain{+8.5} & \textbf{87.6}\gain{+7.6} \\
Qwen3-VL-8B & 76.0 & 70.6 & 62.1 & 53.2 & 96.0 & 89.8 & 78.2 & 94.7 & 78.6 & 86.5 \\
\rowcolor{oursbg}\,+\,\textbf{D\textsuperscript{3}} & \textbf{82.8}\gain{+6.8} & \textbf{78.2}\gain{+7.7} & \textbf{79.4}\gain{+17.3} & \textbf{72.5}\gain{+19.2} & \textbf{96.8}\gain{+0.8} & \textbf{98.9}\gain{+9.1} & \textbf{98.0}\gain{+19.8} & \textbf{98.1}\gain{+3.3} & \textbf{83.2}\gain{+4.6} & \textbf{87.9}\gain{+1.4} \\
Qwen3.5-9B & 79.4 & 66.6 & 60.8 & 56.3 & 96.0 & 81.3 & 61.0 & 97.8 & 75.8 & 87.1 \\
\rowcolor{oursbg}\,+\,\textbf{D\textsuperscript{3}} & \textbf{85.0}\gain{+5.5} & \textbf{80.7}\gain{+14.1} & \textbf{77.8}\gain{+17.0} & \textbf{74.4}\gain{+18.1} & \textbf{96.4}\gain{+0.4} & \textbf{99.5}\gain{+18.2} & \textbf{76.9}\gain{+15.9} & \textbf{98.5}\gain{+0.7} & \textbf{82.5}\gain{+6.7} & \textbf{88.2}\gain{+1.1} \\
InternVL2.5-8B & 73.5 & 69.3 & 63.0 & 52.8 & 94.1 & 99.8 & 89.7 & 95.3 & 70.4 & 78.1 \\
\rowcolor{oursbg}\,+\,\textbf{D\textsuperscript{3}} & \textbf{84.3}\gain{+10.9} & \textbf{81.0}\gain{+11.7} & \textbf{76.9}\gain{+13.9} & \textbf{72.0}\gain{+19.2} & \textbf{95.6}\gain{+1.5} & \textbf{99.9}\gain{+0.1} & \textbf{99.2}\gain{+9.6} & \textbf{98.3}\gain{+3.0} & \textbf{79.1}\gain{+8.7} & \textbf{87.1}\gain{+9.1} \\
InternVL3-8B & 74.7 & 66.6 & 61.6 & 54.4 & 95.1 & 98.0 & 85.6 & 95.1 & 75.9 & 83.4 \\
\rowcolor{oursbg}\,+\,\textbf{D\textsuperscript{3}} & \textbf{82.3}\gain{+7.6} & \textbf{74.5}\gain{+8.0} & \textbf{75.6}\gain{+14.1} & \textbf{68.9}\gain{+14.5} & \textbf{96.1}\gain{+1.0} & \textbf{99.9}\gain{+1.9} & \textbf{98.9}\gain{+13.3} & \textbf{98.4}\gain{+3.2} & \textbf{81.6}\gain{+5.7} & \textbf{87.9}\gain{+4.5} \\
InternVL3.5-8B & 61.8 & 64.7 & 65.7 & 57.0 & 92.2 & 99.2 & 87.3 & 95.1 & 73.5 & 82.8 \\
\rowcolor{oursbg}\,+\,\textbf{D\textsuperscript{3}} & \textbf{81.4}\gain{+19.6} & \textbf{75.2}\gain{+10.4} & \textbf{76.4}\gain{+10.6} & \textbf{70.1}\gain{+13.1} & \textbf{95.1}\gain{+2.8} & \textbf{99.8}\gain{+0.6} & \textbf{98.4}\gain{+11.1} & \textbf{98.4}\gain{+3.3} & \textbf{79.5}\gain{+6.0} & \textbf{87.5}\gain{+4.7} \\
\midrule
Avg. & 72.4 & 68.9 & 61.0 & 54.0 & 91.3 & 94.8 & 78.7 & 90.1 & 69.4 & 79.3 \\
\rowcolor{oursbg}\,+\,\textbf{D\textsuperscript{3}} & \textbf{83.2}\gain{+10.8} & \textbf{77.7}\gain{+8.8} & \textbf{75.1}\gain{+14.1} & \textbf{69.9}\gain{+15.9} & \textbf{93.5}\gain{+2.2} & \textbf{99.7}\gain{+4.9} & \textbf{91.2}\gain{+12.5} & \textbf{97.9}\gain{+7.8} & \textbf{79.4}\gain{+10.0} & \textbf{87.2}\gain{+7.9} \\
\bottomrule
\end{tabular}
\end{adjustbox}
\end{table*}

\subsection{Experimental Setup}
\label{sec:exp_setup}

\paragraph{Benchmarks.}
Our main benchmark is 957 held-out GQA comparison questions~\cite{hudson2019gqa}, yes/no and one per image, on images not used during development.
From 670 of them, the sentence patterns of Sec.~\ref{sec:method} read two entities; these are the color and material questions of Sec.~\ref{sec:findings}.
The other 287 ask about a group of objects, e.g., ``Are these animals of the same species?'', so D\textsuperscript{3} outputs the direct answer on them; accuracy is computed over all 957 questions.
The other benchmarks cover the kinds of comparison in Tab.~\ref{tab:instances}.
For attribute comparisons on other images and annotations, we use those of GQA testdev; two sets built from the human-verified attributes of VAW~\cite{pham2021vaw}, VAW-Nat on whole photos and VAW-Pairs on two cropped objects side by side; VG-Attribution of ARO~\cite{yuksekgonul2023aro}; and those of CLEVR~\cite{johnson2017clevr} (CLEVR-Attr).
For count comparisons we use those of CLEVR (CLEVR-Count); for position, What'sUp~\cite{kamath2023whatsup} and EmbSpatial~\cite{du2024embspatial}; and CV-Bench~\cite{tong2024cambrian} mixes count, position, depth, and distance questions.
All questions have answer options; the Supp.\ describes how each benchmark is built.

\paragraph{Implementation details.}
We evaluate eight frozen MLLMs with 7--9B parameters from three families, each with its official chat template: LLaVA-1.5-7B~\cite{liu2024llava15}; Qwen2-VL-7B~\cite{wang2024qwen2vl}, Qwen2.5-VL-7B~\cite{bai2025qwen25vl}, Qwen3-VL-8B~\cite{bai2025qwen3vl}, and Qwen3.5-9B~\cite{qwen2026qwen35}; and InternVL2.5-8B~\cite{chen2024internvl25}, InternVL3-8B~\cite{zhu2025internvl3}, and InternVL3.5-8B~\cite{wang2025internvl35}.
For attribute and count comparisons, the describer is the model itself; for position, it is Qwen2.5-VL for every model, and a description is a box.
On CV-Bench, Qwen2.5-VL also gives the counts, and geometric tools~\cite{yang2024depthanythingv2,wang2025moge2} describe depth and distance (Sec.~\ref{sec:exp_describers}).
The description-based decision is made by the rule (Tab.~\ref{tab:instances}); on attribute values, the text-only call gives nearly the same accuracy (Sec.~\ref{sec:exp_gain}).
Results for three additional models (LLaVA-OneVision-7B, Pixtral-12B, and InternVL3-14B) are in the Supp.

\paragraph{Evaluation protocol.}
We froze the description prompt, the rule, and the merge weight before any model output on the held-out set, and did not change them afterwards.
Before the evaluation, we also wrote down four predictions for Qwen2.5-VL and InternVL2.5 and fixed the other six models; all four predictions hold (Supp.).
Likewise, on each of the other benchmarks, the sentence patterns and rules were fixed before its test questions were scored, on a development split where there is one.
All methods, including D\textsuperscript{3}, read the answer the same way: from the probabilities the model assigns to the answer options at the first generated token. For baselines that generate free-form answers, such as CoT, we instead extract the answer from the generated text.
For a fair comparison, we report each baseline under the setting most lenient to it: the setting with the highest mean accuracy among those we ran, with the most lenient answer extraction, which raises the baselines' numbers.
For example, with the answer extraction we first specified, CoT reaches 72.1\%, below the direct answer (72.4\%); it reaches 75.2\% once a bare ``yes'' or ``no'' or a closing ``the answer is \dots'' also counts, and we report 75.2\%.
All settings of each baseline and the 95\% confidence interval of each number are in the Supp.

\subsection{Main Results}
\label{sec:exp_main}
Tab.~\ref{tab:main} reports D\textsuperscript{3} on ten benchmarks, and D\textsuperscript{3} improves every model on every benchmark: all 79 model--benchmark pairs gain.
On held-out GQA comparison questions, D\textsuperscript{3} improves the eight models by 2.3--21.5\% with their own descriptions, and their mean accuracy rises from 72.4\% to 83.2\%, higher than that of Qwen2.5-VL-32B answering directly (82.8\%). With InternVL2.5 as the describer, D\textsuperscript{3} improves them by 5.4--28.0\%, to 84.9\% on average. The three additional models also gain (Supp.). With their own descriptions, Qwen2.5-VL and InternVL2.5 gain 6.6\% on average.
The gain also holds on other images and other annotations: 8.8\% on GQA testdev, and 14.1\% and 15.9\% on the two VAW sets; on ARO, where the direct answer is already at 91.3\%, D\textsuperscript{3} still gains 2.2\%.
It holds for the other kinds of description as well: with counts, CLEVR-Count gains 12.5\%; with boxes, What'sUp rises from 90.1\% to 97.9\% and EmbSpatial from 69.4\% to 79.4\%; and with geometric tools added, CV-Bench rises from 79.3\% to 87.2\% (Sec.~\ref{sec:exp_describers}).
The scene annotations of CLEVR also let us check each description: on its attribute comparisons, where accuracy rises from 94.8\% to 99.7\%, both descriptions are right for 96.4\% of the wrong direct answers.
On CLEVR too, as in Finding~1, the model sees the entities and still answers wrongly: the fault lies in the decision.
Can remedies that still let the model decide with the image in view reach the same gain?

\subsection{Comparison with Existing Remedies}
\label{sec:exp_remedies}
If the bias could be fixed while the model still decides with the image in view, some existing remedy should come close to D\textsuperscript{3}.
Tab.~\ref{tab:remedies} compares eight such remedies on held-out GQA: re-prompting, label-free calibration, visual contrastive decoding~\cite{leng2024vcd}, CoT~\cite{wei2022chain}, describing before answering, voting over three phrasings of the question~\cite{wang2023selfconsistency}, fusing the direct answer with that of a second MLLM, and handing back a whole-image caption written by the model, together with the image~\cite{weng2025caption}.
Even at the setting most lenient to each, the best of them, CoT, gains only 2.8\%, answer fusion gains 2.7\%, and the other six change accuracy by between $-$0.5\% and $+$0.6\%; on the same models, D\textsuperscript{3} is at least 6.9\% above every one of them (Supp.).
A reasoning mode does not help either: Qwen3.5 in its thinking mode is 3.1\% below its own direct answer and 8.7\% below D\textsuperscript{3} (Supp.).
Describing before answering shows the reason most directly: the model writes its descriptions and then answers in the same reply, and accuracy hardly changes ($+$0.1\%).
The descriptions are written down, but the decision is still made with the image in view (Finding~4).

Two other remedies decide without the image, as D\textsuperscript{3} does, but from text other than one description per entity: one reads the reasoning text of CoT~\cite{yamaguchi2026red}, and the other a whole-image caption written by the model~\cite{weng2025caption}.
The first gains 3.4\% and stays 7.5\% below D\textsuperscript{3}; the second, even when merged with the direct answer at equal weight as in D\textsuperscript{3}, is 2.9\% below the direct answer and 9.6\% below D\textsuperscript{3}.
The difference lies in the text that is read: even with 512 tokens, a whole-image caption mentions both entities of the question in only 34--51\% of the cases.
Taking the image away is therefore not enough; the decision must also be made from one description per entity.
Finally, as a reference, we fit one threshold for each attribute and wording of the question on the test labels, an upper bound on what calibrating the direct answer can reach: it gives 83.0\%, and D\textsuperscript{3}, which uses no labels, gives 83.2\%.
What such thresholds correct is exactly the preference.
Does the gain of D\textsuperscript{3} also come from removing the preference?

\begin{table}[!t]
\centering
\setlength{\heavyrulewidth}{1.1pt}\setlength{\lightrulewidth}{0.6pt}\setlength{\cmidrulewidth}{0.4pt}
\caption{\textbf{Comparison with existing remedies on held-out GQA.}
Mean accuracy (\%) of the direct answer, the remedy, and D\textsuperscript{3}, over the models each remedy was run on (\#M); subscripts give the gain over the direct answer.
Each remedy is reported at the setting most lenient to it (Sec.~\ref{sec:exp_setup}).
The last row fits one threshold for each attribute and wording of the question on the test labels.
Per-model results are in the Supp.}
\label{tab:remedies}
\renewcommand{\arraystretch}{1.1}
\setlength{\tabcolsep}{3pt}
\begin{adjustbox}{max width=\columnwidth}
\begin{tabular}{@{}l c c c c@{}}
\toprule
Remedy & \#M & Direct & Remedy & \cellcolor{oursbg}\textbf{D\textsuperscript{3}} \\
\midrule
\multicolumn{5}{@{}l}{\textit{Decides with the image in view}} \\
Re-prompting & 6 & 74.0 & 74.4\gain{+0.4} & \cellcolor{oursbg}\textbf{83.9}\gain{+9.9} \\
Label-free calibration & 8 & 72.4 & 72.1\loss{-0.3} & \cellcolor{oursbg}\textbf{83.2}\gain{+10.8} \\
Visual contrastive decoding~\cite{leng2024vcd} & 8 & 72.3 & 71.8\loss{-0.5} & \cellcolor{oursbg}\textbf{83.2}\gain{+10.9} \\
Chain-of-thought~\cite{wei2022chain} & 8 & 72.4 & 75.2\gain{+2.8} & \cellcolor{oursbg}\textbf{83.2}\gain{+10.8} \\
Describe, then answer & 8 & 72.4 & 72.5\gain{+0.1} & \cellcolor{oursbg}\textbf{83.2}\gain{+10.8} \\
Voting over three phrasings~\cite{wang2023selfconsistency} & 8 & 72.4 & 73.0\gain{+0.6} & \cellcolor{oursbg}\textbf{83.2}\gain{+10.8} \\
Answer fusion with a second MLLM & 8 & 72.4 & 75.1\gain{+2.7} & \cellcolor{oursbg}\textbf{83.2}\gain{+10.8} \\
Whole-image caption, with the image~\cite{weng2025caption} & 3 & 77.2 & 76.9\loss{-0.3} & \cellcolor{oursbg}\textbf{83.9}\gain{+6.7} \\
\midrule
\multicolumn{5}{@{}l}{\textit{Decides without the image, from other text}} \\
CoT reasoning~\cite{yamaguchi2026red} & 8 & 72.4 & 75.8\gain{+3.4} & \cellcolor{oursbg}\textbf{83.2}\gain{+10.8} \\
Whole-image caption~\cite{weng2025caption} & 3 & 77.2 & 74.3\loss{-2.9} & \cellcolor{oursbg}\textbf{83.9}\gain{+6.7} \\
\midrule
\textcolor{oraclegray}{Threshold fit on test labels} & 8 & 72.4 & 83.0\gain{+10.6} & \cellcolor{oursbg}\textbf{83.2}\gain{+10.8} \\
\bottomrule
\end{tabular}
\end{adjustbox}
\end{table}

\subsection{Where Does the Gain Come From?}
\label{sec:exp_gain}
\begin{figure}[t]
  \centering
  \includegraphics[width=\linewidth]{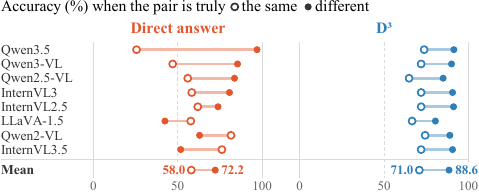}
\caption{\textbf{With D\textsuperscript{3}, the models no longer lean in different directions.}
  Each row is a model, on the 670 questions of Sec.~\ref{sec:findings}.
  Left: with the direct answer, the models lean in different directions.
  Right: with D\textsuperscript{3}, all eight models land at nearly the same place.
  Bottom row: the mean over the models.}
  \label{fig:preference}
\end{figure}

Three quarters of the 670 color and material questions of the held-out set are different pairs (Sec.~\ref{sec:findings}), so a method could gain simply by answering ``different'' more often.
D\textsuperscript{3} does not gain this way.
First, both kinds of questions gain: averaged over the eight models, accuracy rises from 58.0\% to 71.0\% on same pairs and from 72.2\% to 88.6\% on different pairs; over all 957 questions, balanced accuracy (the mean of the accuracies on yes and on no questions) gains 11.6\%, no less than accuracy does (10.8\%).
Second, the models no longer lean in different directions (Fig.~\ref{fig:preference}). With the direct answer, the accuracy gap between different pairs and same pairs ranges from $-$24.4\% to $+$71.4\% across the eight models, and its direction differs from model to model (Finding~2); with D\textsuperscript{3}, the eight models land at nearly the same place, 65--74\% on same pairs and 80--91\% on different pairs. Third, the gain does not depend on the rule: with the model's own text-only call in its place, accuracy is 83.9\%, against 84.1\% with the rule (seven models; Supp.). In total, D\textsuperscript{3} changes 18.2\% of the answers, correcting 1{,}113 and breaking 283 (summed over the eight models).

The gain also does not come from averaging two answers.
With the rule, $p_{\mathrm{desc}}$ is 0 or 1, so the merge simply takes the description-based decision wherever the rule gives one, and the direct answer elsewhere; on held-out GQA, not one of the 7{,}656 answers of the eight models differs from this.
The merge matters when $p_{\mathrm{desc}}$ is a probability instead, as with the text-only call (Fig.~\ref{fig:method}): then the direct answer wins on the questions where it is more confident than the text-only call.
On ARO (six models), for example, the text-only call alone reaches 93.8\%, below the direct answer (94.8\%), but merging the two reaches 96.0\%.
The value of $\lambda$ matters little as long as $\tfrac{1}{2} \le \lambda < 1$ (checked on held-out GQA after the evaluation): with the rule, every such $\lambda$ gives the same answers as $\lambda = \tfrac{1}{2}$, and with the text-only call, accuracy stays between 83.9\% and 84.1\% (seven models).
Below $\tfrac{1}{2}$, the direct answer starts to overrule the description-based decision on the questions where the two disagree and the direct answer is very confident; but it is right on only 30.3\% of these questions at $\lambda = 0.4$, so accuracy drops to 82.1\% (80.1\% at $\lambda = 0.3$).

Where are the remaining errors?
On the 670 questions, the eight models still make 849 errors with D\textsuperscript{3}; in 646 of them the description result itself is wrong, and in 203 the model gives no description.
D\textsuperscript{3} fixes the decision, but not a wrong or missing description.
Can another describer then raise accuracy further?

\subsection{Other Descriptions and Describers}
\label{sec:exp_describers}
Of the three steps of D\textsuperscript{3}, only the first can change: with another describer or another kind of description, the decision is still made without the image and the merge stays the same (Tab.~\ref{tab:instances}).
We first change the describer.
On held-out GQA, when all eight models use the descriptions of InternVL2.5, mean accuracy is 84.9\% (+12.5\%), 1.6\% higher than with the models' own descriptions. In this setting, Qwen2.5-VL rises from 82.2\% to 89.0\%, as we predicted before the evaluation. As its own describer, Qwen2.5-VL answers ``cannot tell'' for 46\% of the entities, so it gains only 2.3\%; with InternVL2.5 as the describer, these questions get a description.
More generally, the choice of describer matters little: when all eight models use the descriptions of any one of six of them, the mean gain is 11.3--12.5\% (Supp.).

We next change the kind of description.
For position questions, a description is a box, and the description-based decision compares the centers of the two boxes.
With its own boxes, Qwen2.5-VL rises from 91.4\% to 97.7\% on What'sUp and from 69.6\% to 78.1\% on EmbSpatial; given to the other seven models, the same boxes raise every one of them on both benchmarks (Tab.~\ref{tab:main}; results with each model's own boxes are in the Supp.).
Finding~4 holds for position too: when the two boxes are handed back and the model answers again with the image in view, the mean accuracy of seven models hardly changes ($-$1.1\% to 0.0\%); only when the decision is made from the two boxes without the image does it gain 6.6--11.7\%.
A position can also be a number: on 661 left-or-right questions on the held-out GQA images, when InternVL2.5 gives each entity's horizontal position from 0 to 100, the eight models rise from 79.5\% to 85.6\% on average.

The describer can also be a geometric tool.
On CV-Bench, a depth estimator~\cite{yang2024depthanythingv2} and a 3D point estimator~\cite{wang2025moge2} describe the entities for the depth and distance questions, and Qwen2.5-VL gives boxes and counts for the others; all eight models gain (Tab.~\ref{tab:main}).
In all these cases only the descriptions change, and the decision made without the image follows them.
Ablations that change one step at a time and the scope of D\textsuperscript{3} are in the Supp.

\section{Related Work}
\label{sec:related}
\paragraph{Answer bias and remedies inside the model.}
Answer bias is an old problem in VQA~\cite{goyal2017making,agrawal2018dont}.
It remains common in MLLMs, whose answers shift with language priors~\cite{guan2024hallusionbench}, the order of the options~\cite{zong2024fool}, and the form of the question~\cite{fan2026tinted}, and sometimes ignore the image completely~\cite{li2024naturalbench}.
Such errors are often blamed on seeing~\cite{tong2024mmvp,kamath2023whatsup}, so most remedies act on the observation side and improve what the model sees~\cite{yang2023som,liu2026sbnb,liu2024pai,chen2025adaptvis}.
Recent analyses find that models often answer wrongly while holding the evidence they need~\cite{fu2025hidden,liu2026sbnb,li2026mllms,nooralahzadeh2026arbitration}.
Other remedies repair the decision inside the model by changing its activations~\cite{golovanevsky2025pixels} or decoding~\cite{leng2024vcd,kim2024code,wang2025mllm}, calibrating the output~\cite{zhang2025debiasing}, CoT prompting~\cite{wei2022chain}, or retraining~\cite{ma2026mechanistic,ttrv}.
But the bias varies from question to question, so a single global correction helps little~\cite{li2024naturalbench}.
In all of these analyses and remedies, the model still decides with the image in view.
We find that on attribute comparisons the bias enters at exactly this step: the model's own description of each entity often already gives the right answer, but the decision it makes with the image in view departs from these descriptions (Sec.~\ref{sec:findings}).
D\textsuperscript{3} therefore does not repair this decision; it decides from the descriptions without the image and merges that decision with the direct answer.

\paragraph{Decomposition and decisions without the image.}
Decomposing a question, perceiving each entity separately, and then comparing the results is a classic idea in neuro-symbolic VQA~\cite{andreas2016nmn,johnson2017inferring,yi2018nsvqa}.
The idea has returned with large models, and the approaches fall into two groups by whether the decider has the image in view.
In the first group, the decision is made without the image: by a generated program~\cite{gupta2023visprog,suris2023vipergpt}, a solver~\cite{dai2026facts}, a text-only language model~\cite{zeng2023socratic,qiao2024prism,li2025vista}, or the MLLM itself once the image is removed~\cite{weng2025caption,kukreja2026dissect,lai2026sycophancy,wang2026solving}; a fixed rule has been used this way to diagnose models~\cite{li2026crisp}.
In the second group, an MLLM decides with the image in view, after reading its own descriptions~\cite{wu2023role,mitra2024ccot,ghosh2025visual} or tool outputs~\cite{li2025visualevidence,vo2026depth,yang2026tvc,hu2024sketchpad}, or by voting or debating over several answers~\cite{wang2023selfconsistency,du2024debate}.
Our results place the bias in this second group: describing before answering does not help on average, and handing the descriptions back with the image in view is clearly worse than deciding from them without the image, in line with earlier studies~\cite{weng2025caption,kukreja2026dissect}.
The model favors descriptions that agree with its direct answer (Sec.~\ref{sec:findings}), the confirmation bias reported for LLMs and MLLMs~\cite{xie2024adaptive,qu2025recot}.
A few recent methods combine the two groups: a decider without the image reads the model's own chain of thought~\cite{yamaguchi2026red}, or a geometric rule decides the spatial questions it is sure about and an MLLM with the image in view decides the rest~\cite{liu2026seer}.
D\textsuperscript{3} returns to the simplest form of the classic idea: the model describes each entity, decides from these descriptions without the image, and merges this decision with its direct answer.
It needs no training, and another MLLM or a geometric tool can also serve as the describer.

\section{Conclusion}
MLLMs often see the two entities of a comparison question correctly and still answer wrongly.
We showed that the answer bias enters not when the model sees, but when it decides with the image in view.
D\textsuperscript{3} therefore lets the model describe each entity with the image in view, decide from these descriptions without the image, and merge this decision with its direct answer.
It needs no training, and all 79 model--benchmark pairs gain.
D\textsuperscript{3} fixes the decision but not a wrong description, so a better describer, such as another MLLM or a geometric tool, can raise accuracy further.

{
    \small
    \bibliographystyle{ieeenat_fullname}
    \bibliography{main}
}

\clearpage
\providecolor{cvprblue}{rgb}{0.21,0.49,0.74}
\providecommand{\mref}[2]{\ifcsname r@#1\endcsname\ref{#1}\else\textcolor{cvprblue}{#2}\fi}
\providecommand{\gain}[1]{\textsubscript{\,\textcolor{gaingreen}{#1}}}
\providecommand{\loss}[1]{\textsubscript{\,\textcolor{oraclegray}{#1}}}
\providecommand{\Dthree}{D\textsuperscript{3}}
\newcommand{\supptab}{\footnotesize\setlength{\tabcolsep}{4pt}\renewcommand{\arraystretch}{1.1}%
\setlength{\heavyrulewidth}{1.0pt}\setlength{\lightrulewidth}{0.5pt}\setlength{\cmidrulewidth}{0.4pt}}
\renewcommand{\topfraction}{0.9}
\renewcommand{\bottomfraction}{0.5}
\renewcommand{\textfraction}{0.1}
\renewcommand{\floatpagefraction}{0.99}
\renewcommand{\dbltopfraction}{0.9}
\renewcommand{\dblfloatpagefraction}{0.99}%
\setcounter{topnumber}{3}
\setcounter{bottomnumber}{2}
\setcounter{totalnumber}{4}
\setcounter{dbltopnumber}{2}
\makeatletter\@secpenalty=0\makeatother
\makeatletter
\renewcommand{\paragraph}{\@startsection{paragraph}{4}{\z@}{0.6ex \@plus .2ex \@minus .1ex}{-1em}{\normalfont\normalsize\bfseries}}
\makeatother
\setlength{\textfloatsep}{9pt plus 2pt minus 2pt}
\setlength{\dbltextfloatsep}{9pt plus 2pt minus 2pt}
\setlength{\floatsep}{7pt plus 2pt minus 2pt}
\setlength{\dblfloatsep}{7pt plus 2pt minus 2pt}
\setlength{\intextsep}{7pt plus 2pt minus 2pt}
\captionsetup[table]{skip=5pt}
\renewcommand{\thetable}{S\arabic{table}}
\renewcommand{\thefigure}{S\arabic{figure}}
\ifdefined\theHtable\renewcommand{\theHtable}{S\arabic{table}}\fi
\ifdefined\theHfigure\renewcommand{\theHfigure}{S\arabic{figure}}\fi
\setcounter{table}{0}
\setcounter{figure}{0}
\twocolumn[{%
\centering
\Large
\textbf{\thetitle}\\
\vspace{0.5em}Supplementary Material \\
\vspace{1.0em}
\normalsize
\includegraphics[width=\textwidth]{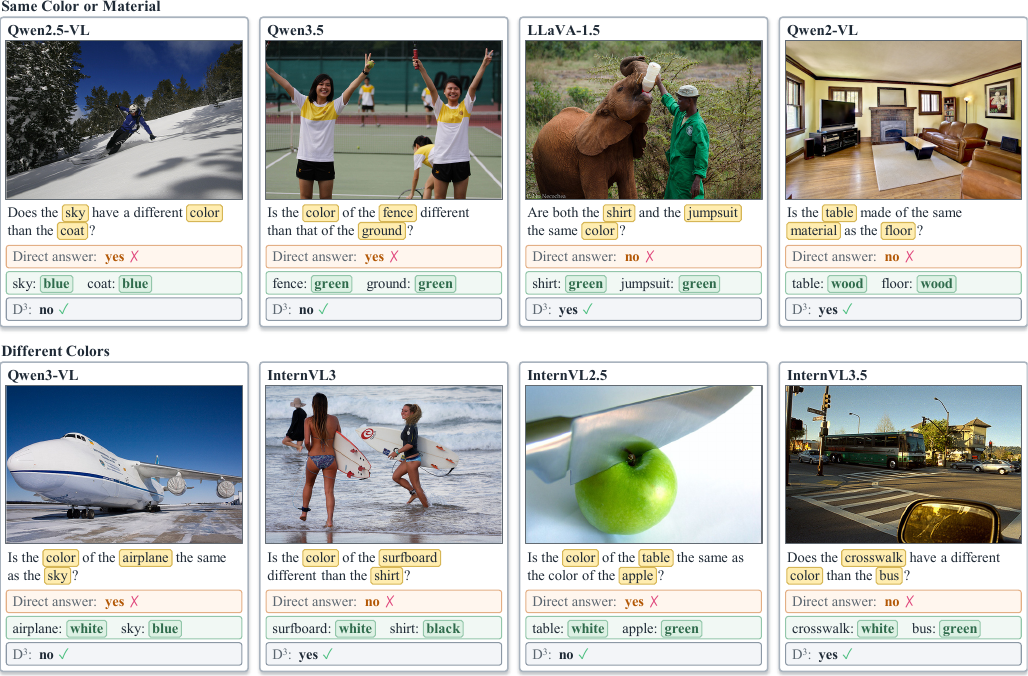}
\captionof{figure}{\textbf{Wrong direct answers that \Dthree{} corrects on held-out GQA, one per model.}
Each card shows the model, the image, the question (entities and aspect in yellow), the direct answer (orange), the model's description of each entity, asked about one entity at a time (green), and the answer of \Dthree{} (gray).
All descriptions match the GQA scene annotation.
Top row: same pairs; bottom row: different pairs.}
\label{fig:supp_examples}
\vspace{1.2em}
}]
\appendix

\noindent This supplementary material is organized as follows.
{\setlength{\topsep}{2pt}\setlength{\partopsep}{0pt}%
\begin{itemize}\setlength{\itemsep}{0pt}\setlength{\parsep}{0pt}%
\item Sec.~\ref{sec:supp_bias}: with \Dthree{}, contradictions between the answers to ``same?'' and ``different?'' fall from 31.4\% to 3.4\%.
\item Sec.~\ref{sec:supp_human}: with blind human labels, the gain is 11.0\%, against 10.5\% with the GQA labels.
\item Sec.~\ref{sec:supp_big}: \Dthree{} raises Qwen2.5-VL-32B, a model four times larger, by 6.1--8.2\% on three benchmarks.
\item Sec.~\ref{sec:supp_entity}: the descriptions are right on natural images.
\item Sec.~\ref{sec:supp_freeform}: on freely reworded questions, \Dthree{} gains 15.1\% without the sentence patterns.
\item Sec.~\ref{sec:supp_gain}: changing one step of \Dthree{} at a time.
\item Sec.~\ref{sec:supp_main}: more models, and each model as the describer.
\item Sec.~\ref{sec:supp_findings}: more on the four findings of Sec.~\mref{sec:findings}{2}.
\item Sec.~\ref{sec:supp_remedies}: every setting of the existing remedies.
\item Sec.~\ref{sec:supp_scope}: the scope of \Dthree{} and the remaining errors.
\item Sec.~\ref{sec:supp_impl}: implementation details.
\item Sec.~\ref{sec:supp_bench}: benchmarks and evaluation protocol.
\end{itemize}}

\section{\Dthree{} Corrects the Answer Bias}
\label{sec:supp_bias}
\begin{table}[!tb]
\centering
\supptab
\caption{\textbf{Two wordings of the same question.} Each of the 670 color and material questions of held-out GQA, asked as ``same?'' and as ``different?'' (\%). Contradict: both answers are ``yes'' or both are ``no''. Both right: both are correct.}
\label{tab:supp_wording}
\setlength{\tabcolsep}{2pt}
\begin{tabularx}{\columnwidth}{@{}l >{\centering\arraybackslash}X >{\centering\arraybackslash}X >{\centering\arraybackslash}X >{\centering\arraybackslash}X@{}}
\toprule
 & \multicolumn{2}{c}{Contradict $\downarrow$} & \multicolumn{2}{c}{Both right $\uparrow$} \\
\cmidrule(lr){2-3}\cmidrule(l){4-5}
Model & Direct & D\textsuperscript{3} & Direct & D\textsuperscript{3} \\
\midrule
LLaVA-1.5 & 91.6 & \cellcolor{oursbg}\textbf{1.0} & 3.1 & \cellcolor{oursbg}\textbf{76.4} \\
Qwen2-VL & 39.4 & \cellcolor{oursbg}\textbf{0.9} & 46.0 & \cellcolor{oursbg}\textbf{84.9} \\
Qwen2.5-VL & 21.2 & \cellcolor{oursbg}\textbf{15.2} & 69.1 & \cellcolor{oursbg}\textbf{74.6} \\
Qwen3-VL & 24.0 & \cellcolor{oursbg}\textbf{3.0} & 59.9 & \cellcolor{oursbg}\textbf{84.2} \\
Qwen3.5 & 5.7 & \cellcolor{oursbg}\textbf{0.6} & 76.7 & \cellcolor{oursbg}\textbf{87.0} \\
InternVL2.5 & 22.4 & \cellcolor{oursbg}\textbf{0.6} & 61.3 & \cellcolor{oursbg}\textbf{86.1} \\
InternVL3 & 23.7 & \cellcolor{oursbg}\textbf{4.8} & 63.0 & \cellcolor{oursbg}\textbf{83.4} \\
InternVL3.5 & 23.1 & \cellcolor{oursbg}\textbf{0.9} & 44.6 & \cellcolor{oursbg}\textbf{85.5} \\
\midrule
Avg. & 31.4 & \cellcolor{oursbg}\textbf{3.4} & 53.0 & \cellcolor{oursbg}\textbf{82.8} \\
\bottomrule
\end{tabularx}
\end{table}

\paragraph{Two wordings of the same question.}
\label{sec:supp_wording}
Finding~2 asks about each pair once as ``same?'' and once as ``different?'', and the two direct answers contradict each other on 31.4\% of the questions.
The describe calls do not see the question, so \Dthree{} uses the same two descriptions for both wordings.
With \Dthree{}, the two answers contradict each other on 3.4\% of the questions, all of them without a description, and both answers are right on 82.8\% instead of 53.0\% (Tab.~\ref{tab:supp_wording}).
Both answers are right more often for every model, on same pairs and on different pairs.
With the text-only call in place of the rule (seven models, as in Finding~4), the contradictions fall from 22.8\% to 4.9\%, and both answers are right on 82.7\% instead of 60.1\%.

\paragraph{Two orders of the options.}
Swapping the two options (``A: no, B: yes'') changes 11.8\% of the direct answers but only 2.7\% of those of \Dthree{} (six models).
The direct answer depends on the form of the question; \Dthree{} answers from the descriptions.

\section{The Gain Holds under Human Labels}
\label{sec:supp_human}
One author, blind to the GQA labels and the model outputs, labeled from the image 120 of the questions whose answer \Dthree{} changes, drawn by a fixed rule.
These labels agree with the GQA labels on 105 of the 120 questions, and all 12 questions shown twice got the same label both times.
Estimated from these questions, the gain on held-out GQA is 11.0\% with the human labels and 10.5\% with their GQA labels.
The labels are available at \url{https://github.com/paul-swu/describe-dont-decide}.

\section{A Larger Model}
\label{sec:supp_big}
\begin{table}[!tb]
\centering
\supptab
\caption{\textbf{A larger model.} Accuracy (\%) of the direct answer and of \Dthree{} with the model's own descriptions, with the gain over the direct answer as subscripts; the 7B numbers are those of Tab.~\mref{tab:main}{2}, and the prompts, rules, and merge those frozen for the eight models.}
\label{tab:supp_big}
\setlength{\tabcolsep}{2pt}
\begin{tabularx}{\columnwidth}{@{}l >{\centering\arraybackslash}X >{\centering\arraybackslash}X >{\centering\arraybackslash}X >{\centering\arraybackslash}X@{}}
\toprule
 & \multicolumn{2}{c}{Qwen2.5-VL-7B} & \multicolumn{2}{c}{Qwen2.5-VL-32B} \\
\cmidrule(lr){2-3}\cmidrule(l){4-5}
Benchmark & Direct & \Dthree{} & Direct & \Dthree{} \\
\midrule
CLEVR-Count & 82.8 & 99.3\gain{+16.5} & 90.2 & 98.2\gain{+8.0} \\
VAW-Nat & 60.5 & 65.9\gain{+5.4} & 66.1 & 72.2\gain{+6.1} \\
VAW-Pairs & 52.9 & 62.7\gain{+9.8} & 59.5 & 67.7\gain{+8.2} \\
\bottomrule
\end{tabularx}
\end{table}

The answer bias does not go away with model size.
We ran \Dthree{} with Qwen2.5-VL-32B, four times larger than the 7B models, as its own describer, with the prompts, rules, and merge frozen for the eight models (Tab.~\ref{tab:supp_big}).
Its direct answer is more accurate than that of Qwen2.5-VL-7B but still wrong on 9.8\% of the CLEVR-Count and 40.5\% of the VAW-Pairs questions.
\Dthree{} raises its accuracy by 8.0\% on CLEVR-Count, 6.1\% on VAW-Nat, and 8.2\% on VAW-Pairs.

\section{Descriptions Are Right on Natural Images}
\label{sec:supp_entity}
\begin{table}[!tb]
\centering
\supptab
\caption{\textbf{Descriptions checked against the GQA scene annotation} (600 questions with annotated entities, summed over the eight models). Share (\%) of the wrong direct answers corrected by \Dthree{} whose two descriptions match the annotation; under each column, the number of corrected answers.}
\label{tab:supp_entity}
\begin{tabular*}{\columnwidth}{@{\extracolsep{\fill}}>{\raggedright\arraybackslash}m{3.0cm} ccc@{}}
\toprule
 & \multicolumn{3}{c}{Corrected by \Dthree{}} \\
\cmidrule(l){2-4}
 & All & Same pairs & Different pairs \\
 & {\scriptsize 1{,}003} & {\scriptsize 280} & {\scriptsize 723} \\
\midrule
Both descriptions as annotated & \textbf{71.6} & \textbf{98.2} & \textbf{61.3} \\
Both as annotated or as $\geq$6 models say & 78.4 & 99.3 & 70.3 \\
\bottomrule
\end{tabular*}
\end{table}

The GQA scene graphs annotate the color and material of each object.
For 600 of the 670 questions of Sec.~\mref{sec:findings}{2}, each entity matches exactly one annotated object, and we count a description as right only if it is the annotated word.
Among the wrong direct answers that \Dthree{} corrects, both descriptions are right for 71.6\%, and for 98.2\% on same pairs (Tab.~\ref{tab:supp_entity}; Fig.~\ref{fig:supp_examples} shows one per model).
Of the descriptions that differ from the annotation, 39.0\% are the word given by at least six of the eight models; counting these as right raises the two shares to 78.4\% and 99.3\%.
On CLEVR (Sec.~\mref{sec:exp_main}{4.2}), both descriptions are right for 96.4\% of the wrong direct answers.

\section{Questions in Free Form}
\label{sec:supp_freeform}
\begin{table}[!tb]
\centering
\supptab
\caption{\textbf{Questions in free form on held-out GQA} (1{,}109 rewrites of 659 questions). Read: share (\%) of the rewrites from which the model reads the two entities and the aspect. Accuracy (\%) of the direct answer and of \Dthree{} on the rewrites, and on the original questions read by the sentence patterns, counted once per rewrite.}
\label{tab:supp_freeform}
\setlength{\tabcolsep}{2pt}
\begin{tabularx}{\columnwidth}{@{}l >{\centering\arraybackslash}X >{\centering\arraybackslash}X >{\centering\arraybackslash}X >{\centering\arraybackslash}X >{\centering\arraybackslash}X@{}}
\toprule
 & \multicolumn{3}{c}{Reworded, read by the model} & \multicolumn{2}{c}{Original, patterns} \\
\cmidrule(lr){2-4}\cmidrule(l){5-6}
Model & Read & Direct & \Dthree{} & Direct & \Dthree{} \\
\midrule
LLaVA-1.5 & 45.9 & 35.6 & 56.0 & 44.8 & 76.7 \\
Qwen2-VL & 99.9 & 64.5 & 84.4 & 67.2 & 85.0 \\
Qwen2.5-VL & 99.8 & 72.9 & 73.5 & 75.1 & 79.0 \\
Qwen3-VL & 99.7 & 72.7 & 84.1 & 75.1 & 84.9 \\
Qwen3.5 & 99.6 & 78.6 & 86.9 & 78.9 & 87.1 \\
InternVL2.5 & 99.2 & 68.4 & 84.2 & 70.1 & 86.1 \\
InternVL3 & 99.7 & 68.7 & 84.6 & 73.9 & 86.0 \\
InternVL3.5 & 99.8 & 57.1 & 85.4 & 57.0 & 85.8 \\
\midrule
\rowcolor{oursbg}[\dimexpr\tabcolsep+0.6pt\relax][\dimexpr\tabcolsep+0.6pt\relax]Avg. & 93.0 & 64.8 & 79.9 & 67.8 & 83.8 \\
\bottomrule
\end{tabularx}
\end{table}

In Tab.~\mref{tab:main}{2}, the entities and the aspect are read from the question with fixed sentence patterns.
To test whether \Dthree{} needs these patterns, Qwen2.5-VL-32B rewrote each of the 670 questions of Sec.~\mref{sec:findings}{2} in three free forms.
A rewrite is kept only if the patterns cannot read it, it names the same two entities, and a check with made-up descriptions gives it the same answer as the original question, which leaves 1{,}109 rewrites of 659 questions.
Each model then reads the two entities, the aspect, and whether the question asks ``same'' or ``different'' from the rewrite itself, in one more call without the image, and the rest of \Dthree{} is unchanged.

On the rewrites, \Dthree{} raises mean accuracy from 64.8\% to 79.9\% (+15.1\%), close to its gain of 16.1\% on the original questions with the patterns (Tab.~\ref{tab:supp_freeform}).
Every model gains, and so do both same pairs (+14.9\%) and different pairs (+15.1\%).
With the text-only call in place of the rule, \Dthree{} reaches 83.9\% (seven models), against 83.3\% with the rule.
Rewording alone lowers the direct answer from 67.8\% to 64.8\%.

\section{What Each Step Contributes}
\label{sec:supp_gain}
\begin{table*}[!t]
\centering
\supptab
\caption{\textbf{Changing one step of \Dthree{} at a time.} Mean accuracy (\%) of the variant and of \Dthree{} over the same \#M models and questions. $^{\dagger}$Both use the text-only call.}
\label{tab:supp_ablations}
\begin{tabular*}{\textwidth}{@{\extracolsep{\fill}}>{\raggedright\arraybackslash}p{6.4cm} l c c c l@{}}
\toprule
Variant & Questions & \#M & Variant & \cellcolor{oursbg}\textbf{\Dthree{}} & Difference \\
\midrule
\multicolumn{6}{@{}l}{\textit{Step 1 changed: how the entities are described}} \\
Describer also given the question & GQA (957) & 6 & 83.2 & \cellcolor{oursbg}\textbf{83.9} & $-$0.7 \\
Describer also given the question & development (1,282) & 2 & 71.5 & \cellcolor{oursbg}\textbf{74.1} & $-$2.6 \\
Both entities in one describe call & same pairs (168) & 8 & 61.8 & \cellcolor{oursbg}\textbf{71.0} & $-$9.2 \\
Both entities in one describe call & mean of same and different pairs & 8 & 76.8 & \cellcolor{oursbg}\textbf{79.8} & $-$3.0 \\
Both entities in one describe call & development (1,282) & 2 & 71.9 & \cellcolor{oursbg}\textbf{74.1} & $-$2.2 \\
\midrule
\multicolumn{6}{@{}l}{\textit{Step 2 changed: how the decision is made}} \\
Decision with the image in view (hand-back) & with both descriptions & 7 & 79.9 & \cellcolor{oursbg}\textbf{88.1} & $-$8.2 \\
Decision with a gray image & with both descriptions & 7 & 86.3 & \cellcolor{oursbg}\textbf{88.1} & $-$1.8 \\
Text-only call instead of the rule & GQA (957) & 7 & 83.9 & \cellcolor{oursbg}\textbf{84.1} & $-$0.2 \\
A soft rule: $\sum_c b_1(c)\,b_2(c)$ over the first-token distributions & GQA (957) & 8 & 82.8 & \cellcolor{oursbg}\textbf{83.2} & $-$0.5 \\
A soft rule & development (1,282) & 8 & 71.7 & \cellcolor{oursbg}\textbf{74.4} & $-$2.7 \\
\midrule
\multicolumn{6}{@{}l}{\textit{Step 3 changed: the merge}} \\
No merge (rule) & GQA (957) & 8 & 83.2 & \cellcolor{oursbg}\textbf{83.2} & +0.0 \\
No merge (text-only call)$^{\dagger}$ & ARO (9,273) & 6 & 93.8 & \cellcolor{oursbg}\textbf{96.0} & $-$2.2 \\
No description-based decision (direct answer) & GQA (957) & 8 & 72.4 & \cellcolor{oursbg}\textbf{83.2} & $-$10.8 \\
\bottomrule
\end{tabular*}
\end{table*}

Tab.~\ref{tab:supp_ablations} changes one step of \Dthree{} and keeps the other two.

\paragraph{Describing each entity alone.}
When both entities are described in one call, they more often get different values.
Accuracy on same pairs then falls from 71.0\% to 61.8\%, for each of the eight models, and the mean over same and different pairs from 79.8\% to 76.8\%.
On the development set, overall accuracy drops by 2.2\% (two models).
When the describer is also given the question, the description result says ``same'' for different pairs more often (7.7\% $\rightarrow$ 9.3\%), and accuracy drops by 0.7\% on the held-out set and 2.6\% on the development set.
\Dthree{} therefore describes each entity alone, without the question.

\paragraph{Deciding without the image.}
Deciding with the image in view costs 8.2\%, and with a gray image in its place only 1.8\%.
When the model's own text-only call makes the description-based decision, as in Fig.~\mref{fig:method}{2}, \Dthree{} reaches 83.9\% on held-out GQA, against 84.1\% with the rule (seven models; LLaVA-1.5 is left out, as in Finding~4), and the call follows the two descriptions 99.7\% of the time.
The results are the same on GQA-testdev (78.4\% against 78.1\%), VAW-Nat (75.6\% against 75.5\%), and ARO (96.0\% for both): without the image, the model decides as the rule does.
A soft rule is 0.5\% lower than the rule on the held-out set and 2.7\% lower on the development set.

\paragraph{The merge.}
With the rule, the merge takes the rule's decision where there is one and the direct answer elsewhere, and every $\lambda \ge \tfrac{1}{2}$ gives the same answers.
With the text-only call, it also lets a more confident direct answer decide, which adds 2.2\% on ARO, and accuracy stays between 83.9\% and 84.1\% for $\tfrac{1}{2} \le \lambda < 1$.
\label{sec:supp_lambda}

\section{More Models and Describers}
\label{sec:supp_main}
\begin{table*}[!t]
\centering
\supptab
\caption{\textbf{Three additional models.} Accuracy (\%) of the direct answer and of \Dthree{}, with the gain over the direct answer as subscripts. Each model uses its own descriptions, except: $^{\S}$attribute words from Qwen2.5-VL; $^{\dagger}$boxes from Qwen2.5-VL; $^{\ddagger}$boxes and counts from Qwen2.5-VL, and depth values and 3D points from geometric tools. ``--'': not evaluated; $^{*}$mean over the two models run.}
\label{tab:supp_additional}
\setlength{\tabcolsep}{2pt}
\begin{tabularx}{\textwidth}{@{}l >{\centering\arraybackslash}X >{\centering\arraybackslash}X >{\centering\arraybackslash}X >{\centering\arraybackslash}X >{\centering\arraybackslash}X >{\centering\arraybackslash}X >{\centering\arraybackslash}X@{}}
\toprule
Model & GQA & ARO$^{\S}$ & \makecell{CLEVR-\\Attr} & \makecell{CLEVR-\\Count} & \makecell{What's-\\Up$^{\dagger}$} & \makecell{Emb-\\Spatial$^{\dagger}$} & \makecell{CV-\\Bench$^{\ddagger}$} \\
 & {\scriptsize 957} & {\scriptsize 9{,}273} & {\scriptsize 1{,}086} & {\scriptsize 1{,}200} & {\scriptsize 4{,}341} & {\scriptsize 3{,}275} & {\scriptsize 2{,}374} \\
\midrule
LLaVA-OneVision-7B & 66.1 & 92.9 & 83.7 & 76.6 & 92.3 & 71.7 & 77.9 \\
\rowcolor{oursbg}[\dimexpr\tabcolsep+0.6pt\relax][\dimexpr\tabcolsep+0.6pt\relax]\,+\,\textbf{\Dthree{}} & \textbf{83.7}\gain{+17.6} & \textbf{95.3}\gain{+2.4} & \textbf{95.2}\gain{+11.5} & \textbf{86.4}\gain{+9.8} & \textbf{98.3}\gain{+6.0} & \textbf{78.7}\gain{+7.0} & \textbf{86.7}\gain{+8.8} \\
Pixtral-12B & 76.7 & 94.3 & 92.4 & -- & 89.9 & 65.9 & 76.1 \\
\rowcolor{oursbg}[\dimexpr\tabcolsep+0.6pt\relax][\dimexpr\tabcolsep+0.6pt\relax]\,+\,\textbf{\Dthree{}} & \textbf{79.1}\gain{+2.4} & \textbf{95.6}\gain{+1.3} & \textbf{99.4}\gain{+6.9} & -- & \textbf{97.3}\gain{+7.4} & \textbf{78.1}\gain{+12.2} & \textbf{86.9}\gain{+10.7} \\
InternVL3-14B & 78.3 & 95.4 & 99.3 & 87.3 & 97.1 & 76.0 & 83.1 \\
\rowcolor{oursbg}[\dimexpr\tabcolsep+0.6pt\relax][\dimexpr\tabcolsep+0.6pt\relax]\,+\,\textbf{\Dthree{}} & \textbf{81.4}\gain{+3.1} & \textbf{96.2}\gain{+0.8} & \textbf{100.0}\gain{+0.7} & \textbf{96.6}\gain{+9.2} & \textbf{98.5}\gain{+1.4} & \textbf{82.0}\gain{+6.0} & \textbf{87.9}\gain{+4.8} \\
\midrule
Avg. & 73.7 & 94.2 & 91.8 & 82.0$^{*}$ & 93.1 & 71.2 & 79.0 \\
\rowcolor{oursbg}[\dimexpr\tabcolsep+0.6pt\relax][\dimexpr\tabcolsep+0.6pt\relax]\,+\,\textbf{\Dthree{}} & \textbf{81.4}\gain{+7.7} & \textbf{95.7}\gain{+1.5} & \textbf{98.2}\gain{+6.4} & \textbf{91.5}\gain{+9.5} & \textbf{98.0}\gain{+4.9} & \textbf{79.6}\gain{+8.4} & \textbf{87.1}\gain{+8.1} \\
\bottomrule
\end{tabularx}
\end{table*}

\begin{table*}[!t]
\centering
\supptab
\caption{\textbf{The choice of describer.} Columns: the model that answers; rows: the describer; underlined: the model describes for itself.}
\label{tab:supp_describers}\label{tab:supp_boxes_b}
\setlength{\tabcolsep}{2pt}
\begin{tabularx}{\textwidth}{@{}l >{\centering\arraybackslash}X >{\centering\arraybackslash}X >{\centering\arraybackslash}X >{\centering\arraybackslash}X >{\centering\arraybackslash}X >{\centering\arraybackslash}X >{\centering\arraybackslash}X >{\centering\arraybackslash}X >{\centering\arraybackslash}X@{}}
\toprule
Describer & \multicolumn{1}{c}{\makebox[0pt][c]{\makecell{LLaVA-\\1.5}}} & \multicolumn{1}{c}{\makebox[0pt][c]{\makecell{Qwen2-\\VL}}} & \multicolumn{1}{c}{\makebox[0pt][c]{\makecell{Qwen2.5-\\VL}}} & \multicolumn{1}{c}{\makebox[0pt][c]{\makecell{Qwen3-\\VL}}} & \multicolumn{1}{c}{\makebox[0pt][c]{\makecell{Qwen3.5\\{}}}} & \multicolumn{1}{c}{\makebox[0pt][c]{\makecell{Intern-\\VL2.5}}} & \multicolumn{1}{c}{\makebox[0pt][c]{\makecell{Intern-\\VL3}}} & \multicolumn{1}{c}{\makebox[0pt][c]{\makecell{Intern-\\VL3.5}}} & Avg. \\
\midrule
\multicolumn{10}{@{}l}{\emph{(a) Held-out GQA: gain (\%) over the direct answer, with the descriptions of the describer in each row}} \\
Direct answer (\%) & 55.9 & 75.8 & 82.2 & 76.0 & 79.4 & 73.5 & 74.7 & 61.8 & 72.4 \\
\midrule
Qwen2-VL & +27.3 & \underline{+12.4} & +6.2 & +6.7 & +4.8 & +10.0 & +7.4 & +19.1 & +11.7 \\
Qwen3-VL & +26.0 & +12.3 & +6.4 & \underline{+6.8} & +5.4 & +9.7 & +8.2 & +18.1 & +11.6 \\
Qwen3.5 & +28.0 & +13.2 & +7.0 & +7.4 & \underline{+5.5} & +10.7 & +8.3 & +19.6 & +12.5 \\
InternVL2.5 & +28.0 & +13.2 & +6.8 & +7.5 & +5.4 & \underline{+10.9} & +8.2 & +20.0 & +12.5 \\
InternVL3 & +25.3 & +12.0 & +6.6 & +6.7 & +5.1 & +9.4 & \underline{+7.6} & +17.6 & +11.3 \\
InternVL3.5 & +27.7 & +12.7 & +6.7 & +7.0 & +5.0 & +10.4 & +7.7 & \underline{+19.6} & +12.1 \\
\midrule
\rowcolor{oursbg}[\dimexpr\tabcolsep+0.6pt\relax][\dimexpr\tabcolsep+0.6pt\relax]Own descriptions (Tab.~\mref{tab:main}{2}) & +21.5 & +12.4 & +2.3 & +6.8 & +5.5 & +10.9 & +7.6 & +19.6 & +10.8 \\
\midrule
\multicolumn{10}{@{}l}{\emph{(b) What'sUp, controlled subset B (308 questions): accuracy (\%), with the boxes of the model in each row}} \\
\midrule
Direct answer & 33.8 & 90.9 & 92.5 & 95.5 & 100.0 & 98.1 & 93.8 & 98.1 & 87.8 \\
\rowcolor{oursbg}[\dimexpr\tabcolsep+0.6pt\relax][\dimexpr\tabcolsep+0.6pt\relax]LLaVA-1.5 & \underline{98.7} & 99.0 & 99.0 & 99.0 & 99.0 & 99.0 & 99.0 & 99.0 & 99.0 \\
\rowcolor{oursbg}[\dimexpr\tabcolsep+0.6pt\relax][\dimexpr\tabcolsep+0.6pt\relax]Qwen2.5-VL & 99.0 & 100.0 & \underline{100.0} & 100.0 & 100.0 & 100.0 & 100.0 & 100.0 & 99.9 \\
\rowcolor{oursbg}[\dimexpr\tabcolsep+0.6pt\relax][\dimexpr\tabcolsep+0.6pt\relax]Qwen3-VL & 100.0 & 100.0 & 100.0 & \underline{100.0} & 100.0 & 100.0 & 100.0 & 100.0 & 100.0 \\
\rowcolor{oursbg}[\dimexpr\tabcolsep+0.6pt\relax][\dimexpr\tabcolsep+0.6pt\relax]Qwen3.5 & 99.7 & 100.0 & 100.0 & 100.0 & \underline{100.0} & 100.0 & 100.0 & 100.0 & 100.0 \\
\rowcolor{oursbg}[\dimexpr\tabcolsep+0.6pt\relax][\dimexpr\tabcolsep+0.6pt\relax]InternVL3 & 99.4 & 99.7 & 99.7 & 99.7 & 99.7 & 99.7 & \underline{99.7} & 99.7 & 99.6 \\
\rowcolor{oursbg}[\dimexpr\tabcolsep+0.6pt\relax][\dimexpr\tabcolsep+0.6pt\relax]InternVL3.5 & 96.4 & 99.7 & 99.7 & 99.7 & 99.7 & 99.7 & 99.7 & \underline{99.7} & 99.3 \\
\bottomrule
\end{tabularx}
\end{table*}

\paragraph{Three additional models.}
\label{sec:supp_additional}
We also ran LLaVA-OneVision-7B, Pixtral-12B, and InternVL3-14B with the prompts, rules, and merge frozen for the eight models (on ARO, with the attribute words of Qwen2.5-VL).
All 20 model--benchmark pairs gain (Tab.~\ref{tab:supp_additional}), on average by 7.7\% on held-out GQA, 6.4\% on CLEVR-Attr, 9.5\% on CLEVR-Count, and 8.4\% on EmbSpatial.
With the descriptions of InternVL2.5, the three models gain 17.9\%, 7.0\%, and 5.9\% on held-out GQA.
All 79 gains in Tab.~\mref{tab:main}{2} are positive, and the 95\% confidence interval of the gain lies above zero for 68 of the 71 cells outside What'sUp.

\paragraph{Each model as the describer.}
\label{sec:supp_describers}
Each of six describers raises every one of the eight models, with a mean gain of 11.3--12.5\% (Tab.~\ref{tab:supp_describers}a).
For six models, the descriptions of another model give nearly the same gain as their own.
Qwen2.5-VL answers ``cannot tell'' for 46.2\% of the entities and decides only 31.6\% of the questions from its own descriptions, and the description result of LLaVA-1.5 is right for only 77.2\%.
With the descriptions of InternVL2.5, \Dthree{} raises Qwen2.5-VL by 6.8\% and LLaVA-1.5 by 28.0\%.
\label{sec:supp_describer_matrix}

\paragraph{Each model's own boxes.}
\label{sec:supp_own_boxes}
In Tab.~\mref{tab:main}{2}, all models use the boxes of Qwen2.5-VL on What'sUp and EmbSpatial.
We also asked LLaVA-1.5, Qwen3-VL, Qwen3.5, InternVL3, and InternVL3.5 for their own boxes on controlled subset B of What'sUp, with the same prompt and rule.
On the questions about front, behind, left, and right (Tab.~\ref{tab:supp_boxes_b}b), each model reaches 98.7--100\% with its own boxes, LLaVA-1.5 from a direct answer of 33.8\%.
The boxes of any of the six models raise the mean of the eight models from 87.8\% to 99.0--100\%.

\section{More on the Four Findings}
\label{sec:supp_findings}
\begin{table}[!tb]
\centering
\supptab
\caption{\textbf{Agreement between two models on held-out GQA} (\%). Above the diagonal: the same direct answer. Below the diagonal (shaded): the same description result.}
\label{tab:supp_agreement}
\setlength{\tabcolsep}{2pt}
\begin{tabularx}{\columnwidth}{@{}l >{\centering\arraybackslash}X >{\centering\arraybackslash}X >{\centering\arraybackslash}X >{\centering\arraybackslash}X >{\centering\arraybackslash}X >{\centering\arraybackslash}X >{\centering\arraybackslash}X >{\centering\arraybackslash}X@{}}
\toprule
 & \rotatebox{90}{LLaVA-1.5} & \rotatebox{90}{Qwen2-VL} & \rotatebox{90}{Qwen2.5-VL} & \rotatebox{90}{Qwen3-VL} & \rotatebox{90}{Qwen3.5} & \rotatebox{90}{InternVL2.5} & \rotatebox{90}{InternVL3} & \rotatebox{90}{InternVL3.5} \\
\midrule
LLaVA-1.5 & -- & 59.0 & 55.2 & 49.8 & 43.5 & 50.6 & 45.8 & 57.5 \\
Qwen2-VL & \cellcolor{oursbg}82.3 & -- & 78.3 & 65.6 & 57.9 & 66.6 & 67.1 & 65.0 \\
Qwen2.5-VL & \cellcolor{oursbg}89.2 & \cellcolor{oursbg}93.4 & -- & 84.8 & 81.6 & 77.8 & 83.9 & 66.0 \\
Qwen3-VL & \cellcolor{oursbg}83.0 & \cellcolor{oursbg}92.4 & \cellcolor{oursbg}93.8 & -- & 83.3 & 76.0 & 75.5 & 61.5 \\
Qwen3.5 & \cellcolor{oursbg}81.8 & \cellcolor{oursbg}91.9 & \cellcolor{oursbg}94.3 & \cellcolor{oursbg}95.4 & -- & 72.4 & 77.1 & 52.4 \\
InternVL2.5 & \cellcolor{oursbg}83.2 & \cellcolor{oursbg}91.0 & \cellcolor{oursbg}94.3 & \cellcolor{oursbg}91.8 & \cellcolor{oursbg}91.6 & -- & 78.2 & 69.1 \\
InternVL3 & \cellcolor{oursbg}84.4 & \cellcolor{oursbg}91.3 & \cellcolor{oursbg}94.8 & \cellcolor{oursbg}94.0 & \cellcolor{oursbg}93.8 & \cellcolor{oursbg}95.5 & -- & 62.1 \\
InternVL3.5 & \cellcolor{oursbg}81.9 & \cellcolor{oursbg}89.8 & \cellcolor{oursbg}96.2 & \cellcolor{oursbg}92.0 & \cellcolor{oursbg}91.2 & \cellcolor{oursbg}92.4 & \cellcolor{oursbg}95.2 & -- \\
\bottomrule
\end{tabularx}
\end{table}

\begin{table}[!tb]
\centering
\supptab
\caption{\textbf{Handing the two descriptions back} (Finding~4; held-out GQA questions with both descriptions) with the image, a uniform gray image, or no image. Follows: answers (\%) that follow the descriptions where they agree or disagree with the direct answer. Not followed: disagreements whose answer does not follow them, summed over the models. Accuracy (\%) with the gain over the direct answer as subscripts.}
\label{tab:supp_handback}
\setlength{\tabcolsep}{2pt}
\begin{tabularx}{\columnwidth}{@{}l >{\centering\arraybackslash}X >{\centering\arraybackslash}X >{\centering\arraybackslash}X >{\centering\arraybackslash}X@{}}
\toprule
Handed back & \multicolumn{2}{c}{Follows} & Not & \\
\cmidrule(lr){2-3}
with & agree & disagree & followed & Accuracy \\
\midrule
\multicolumn{5}{@{}l}{\textit{Own descriptions, seven models, 1,054 disagreements}} \\
the image & 92.2 & 78.7 & 206 & 79.9\gain{+6.8} \\
a gray image & 97.9 & 96.8 & 27 & 86.3\gain{+13.2} \\
\rowcolor{oursbg}[\dimexpr\tabcolsep+0.6pt\relax][\dimexpr\tabcolsep+0.6pt\relax]no image & \textbf{99.7} & \textbf{99.8} & \textbf{2} & \textbf{88.1}\gain{+15.0} \\
\midrule
\multicolumn{5}{@{}l}{\textit{InternVL2.5's descriptions, six models, 1,056 disagreements}} \\
the image & 93.0 & 83.1 & 172 & 81.0\gain{+8.9} \\
\rowcolor{oursbg}[\dimexpr\tabcolsep+0.6pt\relax][\dimexpr\tabcolsep+0.6pt\relax]no image & \textbf{99.7} & \textbf{99.4} & \textbf{6} & \textbf{86.7}\gain{+14.6} \\
\bottomrule
\end{tabularx}
\end{table}

\begin{table*}[!t]
\centering
\supptab\renewcommand{\arraystretch}{1.0}
\caption{\textbf{Every setting of every remedy on held-out GQA.} Mean accuracy (\%) over the \#M models a setting was run on, with the gain over the direct answer as subscripts, and \Dthree{} on the same models. $\star$: the setting in Tab.~\mref{tab:remedies}{3}. Italics: run on two models, or fit on the development set.}
\label{tab:supp_remedies_settings}
\begin{tabular*}{\textwidth}{@{\extracolsep{\fill}}l >{\raggedright\arraybackslash}p{9.9cm} c c c@{}}
\toprule
Remedy & Setting & \#M & Remedy & \cellcolor{oursbg}\textbf{\Dthree{}} \\
\midrule
\multicolumn{5}{@{}l}{\textit{Decides with the image in view}} \\
Re-prompting & both option orders, averaged\,$\star$ & 6 & 74.4\gain{+0.4} & \cellcolor{oursbg}\textbf{83.9}\gain{+9.9} \\
 & asked to check the image again & 6 & 73.9\loss{-0.1} & \cellcolor{oursbg}\textbf{83.9}\gain{+9.9} \\
 & told to answer only from the image & 6 & 73.9\loss{-0.1} & \cellcolor{oursbg}\textbf{83.9}\gain{+9.9} \\
 & options in swapped order & 6 & 73.4\loss{-0.6} & \cellcolor{oursbg}\textbf{83.9}\gain{+9.9} \\
Label-free calibration & contrast with a gray image, $\alpha{=}0.5$\,$\star$ & 8 & 72.1\loss{-0.3} & \cellcolor{oursbg}\textbf{83.2}\gain{+10.8} \\
 & contrast with a gray image, $\alpha{=}1.0$ & 8 & 72.0\loss{-0.4} & \cellcolor{oursbg}\textbf{83.2}\gain{+10.8} \\
 & answer on a gray image subtracted, per question & 8 & 70.5\loss{-1.9} & \cellcolor{oursbg}\textbf{83.2}\gain{+10.8} \\
 & mean answer on gray images subtracted & 8 & 70.3\loss{-2.1} & \cellcolor{oursbg}\textbf{83.2}\gain{+10.8} \\
VCD~\cite{leng2024vcd} & noise on the pixels, $\alpha{=}1.0$ & 8 & 70.8\loss{-1.5} & \cellcolor{oursbg}\textbf{83.2}\gain{+10.9} \\
 & noise on the pixels, $\alpha{=}0.5$ & 8 & 71.5\loss{-0.9} & \cellcolor{oursbg}\textbf{83.2}\gain{+10.9} \\
 & noise in the normalized space, $\alpha{=}1.0$ & 8 & 71.2\loss{-1.1} & \cellcolor{oursbg}\textbf{83.2}\gain{+10.9} \\
 & noise in the normalized space, $\alpha{=}0.5$\,$\star$ & 8 & 71.8\loss{-0.5} & \cellcolor{oursbg}\textbf{83.2}\gain{+10.9} \\
Chain-of-thought~\cite{wei2022chain} & answer line only (first specified) & 8 & 72.1\loss{-0.3} & \cellcolor{oursbg}\textbf{83.2}\gain{+10.8} \\
 & lenient answer extraction\,$\star$ & 8 & 75.2\gain{+2.8} & \cellcolor{oursbg}\textbf{83.2}\gain{+10.8} \\
Describe, then answer & answer line only (first specified) & 8 & 72.3\loss{-0.0} & \cellcolor{oursbg}\textbf{83.2}\gain{+10.8} \\
 & lenient answer extraction\,$\star$ & 8 & 72.5\gain{+0.1} & \cellcolor{oursbg}\textbf{83.2}\gain{+10.8} \\
 & \textit{two calls: a one-sentence description, then the answer with it in the prompt} & 2 & 77.2\loss{-0.7} & \cellcolor{oursbg}\textbf{84.4}\gain{+6.6} \\
Voting~\cite{wang2023selfconsistency} & majority of three direct answers (original, ``same?'', ``different?'')\,$\star$ & 8 & 73.0\gain{+0.6} & \cellcolor{oursbg}\textbf{83.2}\gain{+10.8} \\
Answer fusion & equal weight with the direct answer of InternVL2.5\,$\star$ & 8 & 75.1\gain{+2.7} & \cellcolor{oursbg}\textbf{83.2}\gain{+10.8} \\
 & \textit{Qwen2.5-VL and InternVL2.5, each with the other's direct answer} & 2 & 80.2\gain{+2.3} & \cellcolor{oursbg}\textbf{84.4}\gain{+6.6} \\
Caption, with image~\cite{weng2025caption} & 512-token caption & 3 & 75.7\loss{-1.6} & \cellcolor{oursbg}\textbf{83.9}\gain{+6.7} \\
 & 96-token caption & 3 & 76.0\loss{-1.2} & \cellcolor{oursbg}\textbf{83.9}\gain{+6.7} \\
 & 512-token caption, merged with the direct answer & 3 & 76.9\loss{-0.3} & \cellcolor{oursbg}\textbf{83.9}\gain{+6.7} \\
 & 96-token caption, merged with the direct answer\,$\star$ & 3 & 76.9\loss{-0.3} & \cellcolor{oursbg}\textbf{83.9}\gain{+6.7} \\
\midrule
\multicolumn{5}{@{}l}{\textit{Decides without the image, from other text}} \\
CoT reasoning~\cite{yamaguchi2026red} & text-only call on the reasoning, alone & 8 & 75.7\gain{+3.3} & \cellcolor{oursbg}\textbf{83.2}\gain{+10.8} \\
 & product with the direct answer, exponent 0.1 & 8 & 74.9\gain{+2.5} & \cellcolor{oursbg}\textbf{83.2}\gain{+10.8} \\
 & product with the direct answer, exponent 0.3 & 8 & 75.8\gain{+3.4} & \cellcolor{oursbg}\textbf{83.2}\gain{+10.8} \\
 & product with the direct answer, exponent 0.5 & 8 & 75.6\gain{+3.2} & \cellcolor{oursbg}\textbf{83.2}\gain{+10.8} \\
 & product with the direct answer, exponent 1 & 8 & 75.7\gain{+3.3} & \cellcolor{oursbg}\textbf{83.2}\gain{+10.8} \\
 & product with the direct answer, exponent 10\,$\star$ & 8 & 75.8\gain{+3.4} & \cellcolor{oursbg}\textbf{83.2}\gain{+10.8} \\
Caption~\cite{weng2025caption} & 512-token caption & 3 & 69.8\loss{-7.5} & \cellcolor{oursbg}\textbf{83.9}\gain{+6.7} \\
 & 96-token caption & 3 & 67.8\loss{-9.4} & \cellcolor{oursbg}\textbf{83.9}\gain{+6.7} \\
 & 512-token caption, merged with the direct answer\,$\star$ & 3 & 74.3\loss{-2.9} & \cellcolor{oursbg}\textbf{83.9}\gain{+6.7} \\
 & 96-token caption, merged with the direct answer & 3 & 72.6\loss{-4.7} & \cellcolor{oursbg}\textbf{83.9}\gain{+6.7} \\
\midrule
\multicolumn{5}{@{}l}{\textit{Reference: thresholds fit on labels}} \\
Threshold & one threshold per attribute and wording, fit on the test labels\,$\star$ & 8 & 83.0\gain{+10.6} & \cellcolor{oursbg}\textbf{83.2}\gain{+10.8} \\
 & one threshold for all questions, fit on the test labels & 8 & 74.4\gain{+2.0} & \cellcolor{oursbg}\textbf{83.2}\gain{+10.8} \\
 & \textit{one threshold per attribute and wording, fit on the development set} & 8 & 68.5\loss{-3.9} & \cellcolor{oursbg}\textbf{83.2}\gain{+10.8} \\
 & \textit{one threshold for all questions, fit on the development set} & 8 & 72.1\loss{-0.3} & \cellcolor{oursbg}\textbf{83.2}\gain{+10.8} \\
\bottomrule
\end{tabular*}
\end{table*}

\begin{table*}[!t]
\centering
\supptab
\caption{\textbf{Prompts, the same for all models.} \{question\} is the original question, \{entity\}, \{e1\}, and \{e2\} are the entities read from it, \{v1\} and \{v2\} are their descriptions, \{attribute\} is ``color'' or ``material'', and \{x\} and \{y\} are the two attribute words of the ARO options.}
\label{tab:supp_prompts}
\begin{tabular*}{\textwidth}{@{\extracolsep{\fill}}>{\raggedright\arraybackslash}p{3.0cm} >{\raggedright\arraybackslash}p{13.9cm}@{}}
\toprule
Call & Prompt \\
\midrule
Direct answer (yes/no benchmarks) & Use the image to answer the question.\newline Question: \{question\}\newline A: yes\newline B: no\newline Select the answer supported by the image. Respond with exactly A or B. \\
Direct answer (benchmarks with options) & \{question with its lettered options\}\newline Answer with the option's letter from the given choices directly. \\
Describe: attribute value & Look at the image and answer the question about one object.\newline Question: \{attribute question\}\newline Answer with one or two words, or `cannot tell' if not visible\newline \textit{Attribute question:} ``What is the color of the \{entity\}?'', ``What material is the \{entity\} made of?'' (GQA); ``What is the \{color / material / shape / pattern\} of the \{entity\}?'', ``What pose is the \{entity\} in?'', ``Is the \{entity\} open or closed?'' (VAW; on VAW-Pairs the entity ends with ``on the left / right''); ``What is the \{color / shape / material / size\} of the \{entity\}?'' (CLEVR). \\
Describe: ARO & Which word describes the \{entity\} in the image better: \{x\} or \{y\}? Answer with the word only. \\
Describe: count & Look at the image and count one kind of object.\newline Question: How many \{entity\} are there?\newline Answer with a number only, or `cannot tell' if not visible \\
Describe: box & Give the bounding box of the \{entity\} in the image as four numbers: x1, y1, x2, y2 (left, top, right, bottom). If you cannot tell, say cannot tell. Answer with the four numbers only. \\
Describe: boxes (CV-Bench count) & Give the bounding boxes of all \{entity\} in the image, one box per line, as four numbers x1, y1, x2, y2. If there are none, say none. \\
Describe: horizontal position & Where is the \{entity\} in the image? Answer with one number from 0 to 100 for its horizontal position, where 0 is the left edge and 100 is the right edge. If you cannot tell, say cannot tell. \\
Text-only call (no image) & Two objects in a picture were described as follows: the \{attribute\} of the \{e1\} is \{v1\}; the \{attribute\} of the \{e2\} is \{v2\}.\newline Question: \{question\}\newline A: yes\newline B: no\newline Answer according to the description. Respond with exactly A or B.\newline \textit{On ARO:} ``Two objects in a picture were described as follows: the \{e1\} is \{v1\}; the \{e2\} is \{v2\}.'', then the question and options of the direct answer, with ``image'' replaced by ``picture''. \\
Hand-back with the image (Finding 4) & Use the image to answer the question.\newline Earlier, you described the image as follows: the \{attribute\} of the \{e1\} is \{v1\}; the \{attribute\} of the \{e2\} is \{v2\}.\newline Question: \{question\}\newline A: yes\newline B: no\newline Select the answer supported by the image. Respond with exactly A or B.\newline \textit{Gray-image control:} the same prompt with a uniform gray image of the same size. \\
\bottomrule
\end{tabular*}
\end{table*}

\paragraph{Finding 1: the probability of ``yes'' from two descriptions.}
For the AUROC in Sec.~\mref{sec:findings}{2}, the first-token probabilities of each describe call are normalized over the color or material words that occur at least 20 times among the development descriptions, which gives $b_1(c)$ and $b_2(c)$.
The probability that the two values are equal is $\sum_c b_1(c)\, b_2(c)$; the probability of ``yes'' is this value for ``same?'' and one minus it for ``different?''.

\paragraph{Finding 2: agreement between models.}
\label{sec:supp_agreement}
Two models give the same direct answer on 66.6\% of the held-out questions on average, and on only 43.5\% for one pair.
Their description results agree on 90.8\% on average (at least 81.8\%), more often than their direct answers for all 28 pairs (Tab.~\ref{tab:supp_agreement}).

\paragraph{Finding 3: seeing the entities better.}
\label{sec:supp_finding3}
The 600 questions of Sec.~\ref{sec:supp_entity} have an annotated box for each entity, and the original question is also asked on three new images: the two crops side by side, the image above the two crops, and the image with both boxes drawn.
Each crop enlarges the box by 20\% on each side, and the boxes are drawn as thin white lines; one added line in the prompt names the two entities and says where they are.
For every model, the description result is more accurate than the direct answer on all four images (by 3.7--30.7\% on the original image), while the three new images change mean accuracy by only $-$5.3\% to $+$1.1\%.

\paragraph{Finding 4: handing back the descriptions.}
\label{sec:supp_finding4}
As a control, the hand-back prompt (Tab.~\ref{tab:supp_prompts}) is also run with a uniform gray image of the same size, which keeps the prompt and removes only the content of the image (Tab.~\ref{tab:supp_handback}).
With the image, the model follows its descriptions on 92.2\% of the questions where they agree with its direct answer, but on 78.7\% where they disagree.
With the gray image, these shares are 97.9\% and 96.8\%, and without an image 99.7\% and 99.8\%.
The same holds when both descriptions are the annotated words of Sec.~\ref{sec:supp_entity} (92.8\% and 78.9\% with the image): of the 571 such cases that disagree with the direct answer, the model fails to follow the descriptions in 112 with the image, in 13 with the gray image, and in none without an image.
The descriptions of InternVL2.5 behave in the same way.
Without the image, LLaVA-1.5 follows its own descriptions on only 58.2\% of the questions, and its accuracy (45.7\%) stays at that of its direct answer (46.1\%), although the rule gives 77.2\% on the same descriptions: it cannot compare two values given as text.
Finding~4 and Fig.~\mref{fig:findings}{1}d therefore use the other seven models, on the questions where the model gives both descriptions; in \Dthree{}, the rule decides for \mbox{LLaVA-1.5}.

\section{More on the Existing Remedies}
\label{sec:supp_remedies}
\begin{table*}[!t]
\centering
\supptab
\caption{\textbf{Existing remedies on held-out GQA, per model} (\%), at the setting of Tab.~\mref{tab:remedies}{3}. ``--'': not run. Gray: uses the test labels. Avg.: mean over the \#M models run. $^{\ast}$For InternVL2.5, the second MLLM is itself.}
\label{tab:supp_remedies_per_model}
\setlength{\tabcolsep}{2pt}
\begin{tabularx}{\textwidth}{@{}l >{\centering\arraybackslash}X >{\centering\arraybackslash}X >{\centering\arraybackslash}X >{\centering\arraybackslash}X >{\centering\arraybackslash}X >{\centering\arraybackslash}X >{\centering\arraybackslash}X >{\centering\arraybackslash}X c@{}}
\toprule
Remedy & \multicolumn{1}{c}{\makebox[0pt][c]{\makecell{LLaVA-\\1.5}}} & \multicolumn{1}{c}{\makebox[0pt][c]{\makecell{Qwen2-\\VL}}} & \multicolumn{1}{c}{\makebox[0pt][c]{\makecell{Qwen2.5-\\VL}}} & \multicolumn{1}{c}{\makebox[0pt][c]{\makecell{Qwen3-\\VL}}} & \multicolumn{1}{c}{\makebox[0pt][c]{\makecell{Qwen3.5\\{}}}} & \multicolumn{1}{c}{\makebox[0pt][c]{\makecell{Intern-\\VL2.5}}} & \multicolumn{1}{c}{\makebox[0pt][c]{\makecell{Intern-\\VL3}}} & \multicolumn{1}{c}{\makebox[0pt][c]{\makecell{Intern-\\VL3.5}}} & Avg.\,(\#M) \\
\midrule
\multicolumn{10}{@{}l}{\emph{Decides with the image in view}} \\
Re-prompting & -- & 76.6 & 83.1 & 75.4 & -- & 75.2 & 74.5 & 61.4 & 74.4 {\scriptsize(6)} \\
Label-free calibration & 47.6 & 76.6 & 82.3 & 76.7 & 79.1 & 73.3 & 76.2 & 64.7 & 72.1 {\scriptsize(8)} \\
Visual contrastive decoding~\cite{leng2024vcd} & 54.8 & 74.8 & 82.0 & 76.2 & 79.4 & 73.3 & 73.0 & 61.1 & 71.8 {\scriptsize(8)} \\
Chain-of-thought~\cite{wei2022chain} & 48.4 & 78.4 & 83.7 & 80.4 & 78.8 & 77.0 & 76.3 & 78.7 & 75.2 {\scriptsize(8)} \\
Describe, then answer & 35.3 & 63.6 & 83.1 & 80.5 & 81.7 & 80.6 & 78.3 & 76.8 & 72.5 {\scriptsize(8)} \\
Voting over three phrasings~\cite{wang2023selfconsistency} & 56.0 & 76.1 & 85.0 & 75.1 & 80.2 & 75.4 & 76.4 & 60.0 & 73.0 {\scriptsize(8)} \\
Answer fusion with a second MLLM & 69.1 & 79.0 & 80.2 & 76.4 & 79.3 & 73.5$^{\ast}$ & 75.4 & 68.2 & 75.1 {\scriptsize(8)} \\
Whole-image caption, with the image~\cite{weng2025caption} & -- & -- & 81.1 & 75.8 & -- & 74.0 & -- & -- & 76.9 {\scriptsize(3)} \\
\multicolumn{10}{@{}l}{\emph{Decides without the image, from other text}} \\
CoT reasoning~\cite{yamaguchi2026red} & 52.8 & 75.9 & 83.6 & 80.9 & 81.2 & 77.1 & 76.0 & 78.8 & 75.8 {\scriptsize(8)} \\
Whole-image caption~\cite{weng2025caption} & -- & -- & 76.5 & 73.6 & -- & 72.8 & -- & -- & 74.3 {\scriptsize(3)} \\
\midrule
\textcolor{oraclegray}{Threshold fit on test labels} & \textcolor{oraclegray}{80.4} & \textcolor{oraclegray}{87.2} & \textcolor{oraclegray}{87.2} & \textcolor{oraclegray}{80.5} & \textcolor{oraclegray}{82.0} & \textcolor{oraclegray}{82.9} & \textcolor{oraclegray}{83.3} & \textcolor{oraclegray}{80.8} & \textcolor{oraclegray}{83.0 {\scriptsize(8)}} \\
\midrule
Direct answer & 55.9 & 75.8 & 82.2 & 76.0 & 79.4 & 73.5 & 74.7 & 61.8 & 72.4 \\
\rowcolor{oursbg}[\dimexpr\tabcolsep+0.6pt\relax][\dimexpr\tabcolsep+0.6pt\relax]\textbf{\Dthree{}} & \textbf{77.4} & \textbf{88.2} & 84.5 & \textbf{82.8} & \textbf{85.0} & \textbf{84.3} & \textbf{82.3} & \textbf{81.4} & \textbf{83.2} \\
\bottomrule
\end{tabularx}
\end{table*}

\begin{table*}[!t]
\centering
\supptab
\caption{\textbf{How each benchmark is built.} \#Q: test questions. Development: the split used before the test questions were scored, and what was fixed on it. Describer: who describes each entity in Tab.~\mref{tab:main}{2}. The last row is the left-or-right set of Sec.~\mref{sec:exp_describers}{4.5}.}
\label{tab:supp_benchmarks}
\begin{tabular*}{\textwidth}{@{\extracolsep{\fill}}l r >{\raggedright\arraybackslash}p{5.3cm} >{\raggedright\arraybackslash}p{3.4cm} l >{\raggedright\arraybackslash}p{2.6cm}@{}}
\toprule
Benchmark & \#Q & Questions & Development: fixed on it & Answers & Describer \\
\midrule
GQA (held-out) & 957 & GQA \texttt{val\_all}: every \emph{compare} yes/no question, one per unused image & 1{,}282 questions: prompt, rule, $\lambda$ & 44.0\% yes & the model: a value \\
GQA-testdev & 326 & GQA \texttt{testdev\_balanced}: color and material comparisons of two named objects & 80 other images: patterns & 41.7\% yes & the model: a value \\
VAW-Nat & 554 & VAW: two objects in one Visual Genome photo; half same pairs & none & 50.0\% yes & the model: a value \\
VAW-Pairs & 1{,}200 & VAW: two objects cropped from two images, side by side; six attributes & none & 50.0\% yes & the model: a value \\
ARO & 9{,}273 & VG-Attribution: the 158 attribute pairs of the official score & 1{,}796 items: rule & 50.6\% first & the model: one of two words \\
CLEVR-Attr & 1{,}086 & CLEVR val.: equality of color, shape, material, or size & 269 questions: patterns & 48.1\% yes & the model: a value \\
CLEVR-Count & 1{,}200 & CLEVR val.: comparison of two counts & 300 images: patterns & 47.8\% yes & the model: a count \\
What'sUp & 4{,}341 & all six subsets (controlled; COCO and VG, one or two objects) & 204 + 413 questions: box rule, 2\% margin & 4 or 2 options & Qwen2.5-VL: a box \\
EmbSpatial & 3{,}275 & EmbSpatial-Bench: position and closest / farthest & 365 questions: box reading & 4 options & Qwen2.5-VL: a box \\
CV-Bench & 2{,}374 & count, relation, depth, distance; official score & 264 questions: describer per type & 2--6 options & Qwen2.5-VL: boxes; depth and 3D tools \\
\midrule
GQA left / right & 661 & GQA \texttt{val\_all} on the held-out images: left / right questions & 700 questions: describer, no margin & 50.1\% yes & InternVL2.5: a position \\
\bottomrule
\end{tabular*}
\end{table*}

\paragraph{Settings.}
All remedies use the direct-answer prompt and run on the 957 held-out GQA questions, and each prompt was fixed before the remedy was run.
Re-prompting swaps the two options, averages over both orders, or adds one sentence (``Double-check the image carefully before you answer.'' or ``Answer only from what is visible in this image, not from what is typical or expected.'').
Label-free calibration contrasts the answer with the answer for a uniform gray image, and visual contrastive decoding (VCD)~\cite{leng2024vcd} with the answer for a noised copy of the image.
Chain-of-thought~\cite{wei2022chain} asks the model to think step by step, and describe-then-answer to first describe each object in one sentence; both end with ``Answer: A'' or ``Answer: B'' and generate at most 256 tokens.
Voting~\cite{wang2023selfconsistency} takes the majority of the direct answers to the question and to its ``same?'' and ``different?'' forms.
Answer fusion merges the direct answers of the model and of InternVL2.5 at equal weight.
The whole-image caption~\cite{weng2025caption} answers ``Describe this image in detail.'' and is handed back with the image or read by the text-only call.
CoT reasoning~\cite{yamaguchi2026red} gives the reasoning of the chain-of-thought reply to the text-only call.
The thresholds are fit for each model, attribute, and wording on the test labels.

\paragraph{Results.}
No setting that decides with the image in view gains more than 2.8\% (Tab.~\ref{tab:supp_remedies_settings}), and on the same models \Dthree{} is at least 6.9\% above every remedy that uses no labels (Tab.~\ref{tab:supp_remedies_per_model}).

\paragraph{Thinking mode.}
\label{sec:supp_thinking}
With its thinking mode on (at most 2{,}048 new tokens), Qwen3.5 is 3.1\% below its own direct answer and 8.7\% below \Dthree{} on held-out GQA, while it generates 597 tokens per question.

\paragraph{What does a threshold correct?}
\label{sec:supp_threshold}
The thresholds in Tab.~\mref{tab:remedies}{3} reach 83.0\% by answering ``different'' on almost every question; answering ``different'' on every question already gives 76.8\%.
On different pairs they reach 97.6\%, but on same pairs they fall from 58.0\% to 31.1\% (Tab.~\ref{tab:supp_threshold}).
\Dthree{} raises both, to 88.6\% and 71.0\%, and its mean over same and different pairs is 79.8\%, against 64.3\% for the thresholds.
Fit on the development set, the thresholds reach only 68.5\% on the held-out set, below the direct answer (72.4\%).
No other remedy raises accuracy on same pairs by more than 2.3\%, while \Dthree{} raises it by 13.0\%.

\raggedbottom
\section{Scope and Remaining Errors}
\label{sec:supp_scope}
\begin{table*}[t]
\centering
\supptab
\caption{\textbf{The four predictions written down before the held-out evaluation.} Accuracy (\%) of the reference and of \Dthree{} on held-out GQA (\#Q: all 957 questions, or the 670 color and material questions). $p$ is the chance of a difference at least this large if \Dthree{} and the reference were right equally often; for predictions 2--4, it is corrected for testing three.}
\label{tab:supp_predictions}
\begin{tabular*}{\textwidth}{@{\extracolsep{\fill}}c >{\raggedright\arraybackslash}p{6.7cm} >{\raggedright\arraybackslash}p{3.1cm} r c l c@{}}
\toprule
 & \Dthree{} & Reference & \#Q & Accuracy & Difference & $p$ \\
\midrule
1 & Qwen2.5-VL with InternVL2.5's descriptions & Qwen2.5-VL, direct & 957 & 82.2\,$\rightarrow$\,89.0 & +6.8 & $2\times10^{-5}$ \\
2 & Qwen2.5-VL and InternVL2.5, own descriptions (mean) & their direct answers & 957 & 77.8\,$\rightarrow$\,84.4 & +6.6 & $6\times10^{-5}$ \\
3 & Qwen2.5-VL with InternVL2.5's descriptions & InternVL2.5, direct & 670 & 70.9\,$\rightarrow$\,86.4 & +15.5 & $6\times10^{-5}$ \\
4 & Qwen2.5-VL with InternVL2.5's descriptions & both direct answers merged & 670 & 77.5\,$\rightarrow$\,86.4 & +9.0 & $6\times10^{-5}$ \\
\bottomrule
\end{tabular*}
\end{table*}

\begin{table}[!tb]
\centering
\supptab
\caption{\textbf{Sentence patterns on GQA.} \{A\} and \{B\} are the two entities, \{same\} is ``same'' or ``different'', and \{aspect\} is ``color(s)'' or ``material(s)''; words in brackets are optional. \#: held-out questions read by the pattern.}
\label{tab:supp_patterns}
\begin{tabular*}{\columnwidth}{@{\extracolsep{\fill}}>{\raggedright\arraybackslash}p{7.5cm} r@{}}
\toprule
Pattern & \# \\
\midrule
Is / are [both] the \{A\} and the \{B\} [the / a] \{same\} \{aspect\}? & 86 \\
Is / are [both] the \{A\} and the \{B\} made [out] of [the / a] \{same\} material? & 19 \\
Do / does [both] the \{A\} and the \{B\} have [the / a] \{same\} \{aspect\}? & 193 \\
Do / does the \{A\} have [the / a] \{same\} \{aspect\} as / than the \{B\}? & 131 \\
Is / are the \{A\} [the / a] \{same\} \{aspect\} as / than the \{B\}? & 40 \\
Is / are the \{A\} made [out] of [the / a] \{same\} material as / than the \{B\}? & 15 \\
Is / are the \{A\} \{same\} in \{aspect\} as / than the \{B\}? & 62 \\
Is / are the \{aspect\} of the \{A\} [the] \{same\} as / than the \{B\}? & 124 \\
\bottomrule
\end{tabular*}
\end{table}

\begin{table}[!tb]
\centering
\supptab
\caption{\textbf{Cost per question on held-out GQA,} averaged over the models a method was run on (\#M). Tokens: generated tokens. Input and Time: prompt tokens and generation time relative to the direct answer.}
\label{tab:supp_cost}
\setlength{\tabcolsep}{2pt}
\begin{tabularx}{\columnwidth}{@{}l >{\centering\arraybackslash}X >{\centering\arraybackslash}X >{\centering\arraybackslash}X >{\centering\arraybackslash}X >{\centering\arraybackslash}X@{}}
\toprule
Method & \#M & Calls & Tokens & Input & Time \\
\midrule
Direct answer & 8 & 1.00 & 1.0 & 1.0$\times$ & 1.0$\times$ \\
\rowcolor{oursbg}[\dimexpr\tabcolsep+0.6pt\relax][\dimexpr\tabcolsep+0.6pt\relax]\Dthree{}, rule & 8 & 2.40 & 4.3 & 2.4$\times$ & 2.8$\times$ \\
\rowcolor{oursbg}[\dimexpr\tabcolsep+0.6pt\relax][\dimexpr\tabcolsep+0.6pt\relax]\Dthree{}, text-only call & 8 & 3.01 & 4.9 & 2.5$\times$ & 3.1$\times$ \\
\midrule
Re-prompting & 6 & 2.00 & 2.0 & 2.0$\times$ & 2.0$\times$ \\
Label-free calib. & 8 & 2.00 & 2.0 & 2.0$\times$ & 2.0$\times$ \\
VCD & 8 & 2.00 & 2.0 & 2.0$\times$ & 2.0$\times$ \\
Chain-of-thought & 8 & 1.00 & 57.6 & 1.0$\times$ & 21.7$\times$ \\
Describe, then answer & 8 & 1.00 & 28.4 & 1.0$\times$ & 10.1$\times$ \\
Voting & 8 & 2.40 & 2.4 & 2.4$\times$ & 2.4$\times$ \\
Answer fusion & 8 & 1.88 & 1.9 & 5.8$\times$ & 6.7$\times$ \\
Caption, with image & 3 & 3.00 & 97.9 & 3.2$\times$ & 36.2$\times$ \\
\midrule
CoT reasoning & 8 & 2.82 & 59.4 & 2.3$\times$ & 23.2$\times$ \\
Caption, no image & 3 & 3.00 & 253.4 & 2.8$\times$ & 110.8$\times$ \\
\bottomrule
\end{tabularx}
\end{table}

\begin{table}[!tb]
\centering
\supptab
\caption{\textbf{Accuracy on same and different pairs: thresholds fit on the test labels and \Dthree{}.} Mean accuracy (\%) of the eight models on held-out GQA. Mean: over same and different pairs. The thresholds are fit for each attribute and wording, as in Tab.~\mref{tab:remedies}{3}.}
\label{tab:supp_threshold}
\setlength{\tabcolsep}{2pt}
\begin{tabularx}{\columnwidth}{@{}l >{\centering\arraybackslash}X >{\centering\arraybackslash}X c >{\centering\arraybackslash}X@{}}
\toprule
 & All & Same & Different & Mean \\
 & {\scriptsize 957} & {\scriptsize 168} & {\scriptsize 502} & \\
\midrule
Direct answer & 72.4 & 58.0 & 72.2 & 65.1 \\
Always ``different'' & 76.8 & 0.0 & 100.0 & 50.0 \\
Thresholds, development set & 68.5 & 67.4 & 64.1 & 65.8 \\
Thresholds, test labels & 83.0 & 31.1 & 97.6 & 64.3 \\
\rowcolor{oursbg}[\dimexpr\tabcolsep+0.6pt\relax][\dimexpr\tabcolsep+0.6pt\relax]\textbf{\Dthree{}} & \textbf{83.2} & \textbf{71.0} & \textbf{88.6} & \textbf{79.8} \\
\bottomrule
\end{tabularx}
\end{table}

\Dthree{} acts on a question when its entities and aspect can be read from it and the answer follows from one description per entity.
This covers the comparisons of Tab.~\mref{tab:instances}{1}; every other question keeps its direct answer.
On all 10{,}119 GQA testdev questions of the 318 test images, \Dthree{} changes only the answers to attribute comparisons, which gain 8.8\%, and overall accuracy rises for each of the eight models.

Of the 849 errors that remain on the 670 color and material questions (Sec.~\mref{sec:exp_gain}{4.4}), all come from the describe step: 646 from a wrong description result and 203 from a missing description, 118 of them from Qwen2.5-VL.

\section{Implementation Details}
\label{sec:supp_impl}

\paragraph{Prompts.}
Tab.~\ref{tab:supp_prompts} lists all prompts, the same for all models.
Each call starts a new conversation in the model's chat template and decodes greedily.
A describe call generates at most 8 new tokens (32 for boxes and positions, and for Qwen2.5-VL and InternVL2.5 on ARO).
The direct answer and the text-only call generate one token; $p_{\mathrm{img}}$ and $p_{\mathrm{desc}}$ are the softmax of the logits of the option letters.

\paragraph{Reading the entities and the aspect.}
The entities and the aspect are read from the question with fixed sentence patterns, before any model call.
On GQA, the eight patterns of Tab.~\ref{tab:supp_patterns}, written from the development questions, read 670 of the 957 held-out questions.
A question is read only if it has exactly one reading and its two entities differ.
Other benchmarks come from templates, so their patterns follow the template text and read all questions of the kinds in Tab.~\mref{tab:instances}{1}.

\paragraph{Descriptions and rules.}
A reply is lower-cased, and quotes, final punctuation, and a leading article are removed.
A reply that contains ``cannot tell'' gives no description.
A box is read as the first four numbers of the reply, a count as its first integer, and a position as its first number between 0 and 100.
For attribute values, the rule answers ``same'' if the two values are equal and ``different'' otherwise, without a list of synonyms; on ARO, it decides only when the two chosen words match one of the options.
For counts and positions, it compares the two numbers; for boxes, the centers of the two boxes along the axis of the relation, with no decision when they are closer than 2\% of the image width.
This margin was set on the development questions of What'sUp and is the only threshold in the rules.

\paragraph{One or three entities.}
Two subsets of What'sUp (3{,}066 questions) ask about one object and the image, and the rule compares the center of the object's box with that of the image.
The count questions of CV-Bench (709) name one kind of object, counted as the number of boxes the describer gives.
The distance questions of CV-Bench (540) name an anchor and two candidates, each entity gets a 3D point, and the rule picks the candidate closer to the anchor.
As with two entities, each is described alone, without the question.

\paragraph{Geometric tools.}
On the depth and distance questions of CV-Bench, Depth Anything V2 (Small)~\cite{yang2024depthanythingv2} describes each entity by the mean relative inverse depth of its marked region, and MoGe-2 (ViT-L)~\cite{wang2025moge2} by its median 3D point.

\paragraph{How the merge weight was set.}
$\lambda = \tfrac{1}{2}$ was our default from the first experiments on the development set.
A later grid search on the development set kept it, and it was frozen with the prompts and rules before any held-out output.

\paragraph{Models and cost.}
All models run from Hugging Face checkpoints in bfloat16, with their own chat templates and image processors, Qwen3.5 with its thinking mode off.
On held-out GQA, \Dthree{} makes 2.40 calls, generates 4.3 tokens per question, and takes 2.8 times the generation time of the direct answer (Tab.~\ref{tab:supp_cost}); chain-of-thought takes 21.7 times.

\section{Benchmarks and Evaluation Protocol}
\label{sec:supp_bench}

\paragraph{Benchmarks.}
Tab.~\ref{tab:supp_benchmarks} summarizes how each benchmark of Tab.~\mref{tab:main}{2} is built.
Where a benchmark has a development split, the sentence patterns and rules were fixed on it before the test questions were scored, and all results are on the test questions.
Before the test questions, each model answers a few check questions, and a model whose replies are not in the expected form is not run on that benchmark.
This is why LLaVA-1.5 has no result on CLEVR-Attr in Tab.~\mref{tab:main}{2}: 7 of its 16 check descriptions were not CLEVR attribute words, such as ``plastic'' and ``gold''.

\paragraph{Predictions.}
Before any model output on held-out GQA existed, we froze the description prompt, the rule, the merge weight, the questions, and the calls, and wrote down four predictions with their tests (Tab.~\ref{tab:supp_predictions}); all four hold.
The same protocol fixed the other six models, run with the same frozen prompts, rule, and merge.

\paragraph{Development set.}
On the 1{,}282 development questions, \Dthree{} improves all eight models with their own descriptions, by 2.4--12.3\% (mean 66.0\% $\rightarrow$ 74.4\%).
Findings~1 and~2 were first seen on this set: the description result is more accurate than the direct answer for every model, and each model leans the same way as on the held-out set, where all eight gain again and the mean gain rises from 8.4\% to 10.8\%.

\end{document}